\documentclass[journal,twoside]{IEEEtran}

\usepackage[table,dvipsnames, svgnames, x11names]{xcolor}
\usepackage{graphicx}
\usepackage{amsfonts,amssymb,amsmath,bm,textcomp}
\usepackage{array}
\usepackage[singlelinecheck=false, labelsep=period, font=footnotesize]{caption}
\usepackage{nicematrix}
\usepackage[ruled,vlined]{algorithm2e}
\usepackage{mathrsfs,tabularx,multirow,makecell,setspace,subfig,color,booktabs,multirow,diagbox,xspace}
\usepackage{float}
\usepackage{bm}
\usepackage[american]{babel}
\usepackage[colorlinks, citecolor=black, bookmarks=true, CJKbookmarks=true, urlcolor=black, bookmarksnumbered=true, bookmarksopen=true]{hyperref}
\usepackage{balance}
\usepackage[utf8]{inputenc}
\usepackage{tikz}
\usepackage{bbm}
\usepackage[capitalize]{cleveref}
\usetikzlibrary{spy}

\newcommand{\squishlist}{
	\begin{list}{$\bullet$}
		{ \setlength{\itemsep}{0pt}
			\setlength{\parsep}{2pt}
			\setlength{\topsep}{2pt}
			\setlength{\partopsep}{0pt}
			\setlength{\leftmargin}{1em}
			\setlength{\labelwidth}{1em}
			\setlength{\labelsep}{0.5em} } }
	\newcommand{\squishend}{
\end{list} }

\newcommand{\ldbracket}{{[\kern-0.17em[}}
\newcommand{\rdbracket}{{]\kern-0.17em]}}

\usepackage{fancyhdr}

\def\ourmethod{{EasyAIF}\xspace}

\begin{document}
	
	\title{Point-and-Shoot All-in-Focus Photo Synthesis \mbox{from Smartphone Camera Pair}
	}
	
	\author{Xianrui~Luo,~Juewen~Peng,~Weiyue~Zhao,~Ke~Xian, Hao~Lu,~and~Zhiguo~Cao

		\thanks{
			%Manuscript received XXXX XX, XXXX; revised XXXX XX, XXXX; accepted XXXX XX, XXXX. Date of publication XXXX XX, XXXX; date of current version XXXX XX, XXXX. 
            This work was supported in part by the National Natural Science Foundation of China (Grant No. U1913602) and funded by Huawei Technologies CO., LTD.
			\textit{(Corresponding author: Zhiguo Cao)}.
			
			Xianrui Luo, Juewen Peng, Weiyue Zhao, Hao Lu, and Zhiguo Cao are with the Key Laboratory of Image Processing and Intelligent Control, Ministry of Education, and also with the School of Artificial Intelligence and Automation, Huazhong University of Science and Technology, Wuhan 430074, China (e-mail: xianruiluo@hust.edu.cn; juewenpeng@hust.edu.cn; 
			zhaoweiyue@hust.edu.cn;
			hlu@hust.edu.cn;
			zgcao@hust.edu.cn).
			
			Ke Xian is with the S-Lab for Advanced Intelligence, Nanyang Technological University, Singapore (e-mail:ke.xian@ntu.edu.cg).
			
			% Color versions of one or more of the figures in this article are available online at http://ieeexplore.ieee.org.
			
            Digital Object Identifier 10.1109/TCSVT.2022.3222609
		}%
	}
	
	\markboth{Manuscript Submitted to IEEE Trans. on Circuit Syst. Video Technol.}%
	{Point-and-Shoot All-in-Focus Photo Synthesis from Smartphone Camera Pair}
	
	\maketitle
	
    \thispagestyle{fancy}
    \fancyhead{}
    \lhead{}
    \lfoot{}
    \cfoot{\small{Copyright \copyright~2022 IEEE. Personal use is permitted, but republication/redistribution requires IEEE permission.\\See \url{http://www.ieee.org/publications_standards/publications/rights/index.html} for more information.}}
    \rfoot{}
	\begin{abstract}
	All-in-Focus (AIF) photography is expected to 
be a commercial selling point for modern smartphones.
Standard AIF synthesis requires manual, time-consuming operations such as focal stack compositing, which is unfriendly to ordinary people. To achieve point-and-shoot AIF photography with a smartphone, we expect that an AIF photo can be generated from one shot of the scene, instead of from multiple photos captured by the same camera. Benefiting from the multi-camera module in modern smartphones, we introduce a new task of AIF synthesis from main (wide) and ultra-wide cameras. 
The goal is to recover sharp details from defocused regions in the main-camera photo with the help of the ultra-wide-camera one.
The camera setting poses new challenges such as parallax-induced occlusions and inconsistent color between cameras. 
To overcome the challenges, we introduce a predict-and-refine network to mitigate occlusions and propose dynamic frequency-domain alignment for color correction. 
To enable effective training and evaluation, we also build an AIF dataset with $2686$ unique scenes. Each scene includes two photos captured by the main camera, one photo captured by the ultra-wide camera, and a synthesized AIF photo.
Results show that our solution, termed \ourmethod, can produce high-quality AIF photos and outperforms strong baselines quantitatively and qualitatively. 
For the first time, we demonstrate point-and-shoot AIF photo synthesis successfully from main and ultra-wide cameras.

	\end{abstract}

	\begin{IEEEkeywords}
		All-in-Focus synthesis, main/ultra-wide camera, occlusion-aware networks
	\end{IEEEkeywords}

	\begin{figure*}
        % 	\vspace{-0.5cm}
            \centering
            \includegraphics[width=\textwidth]{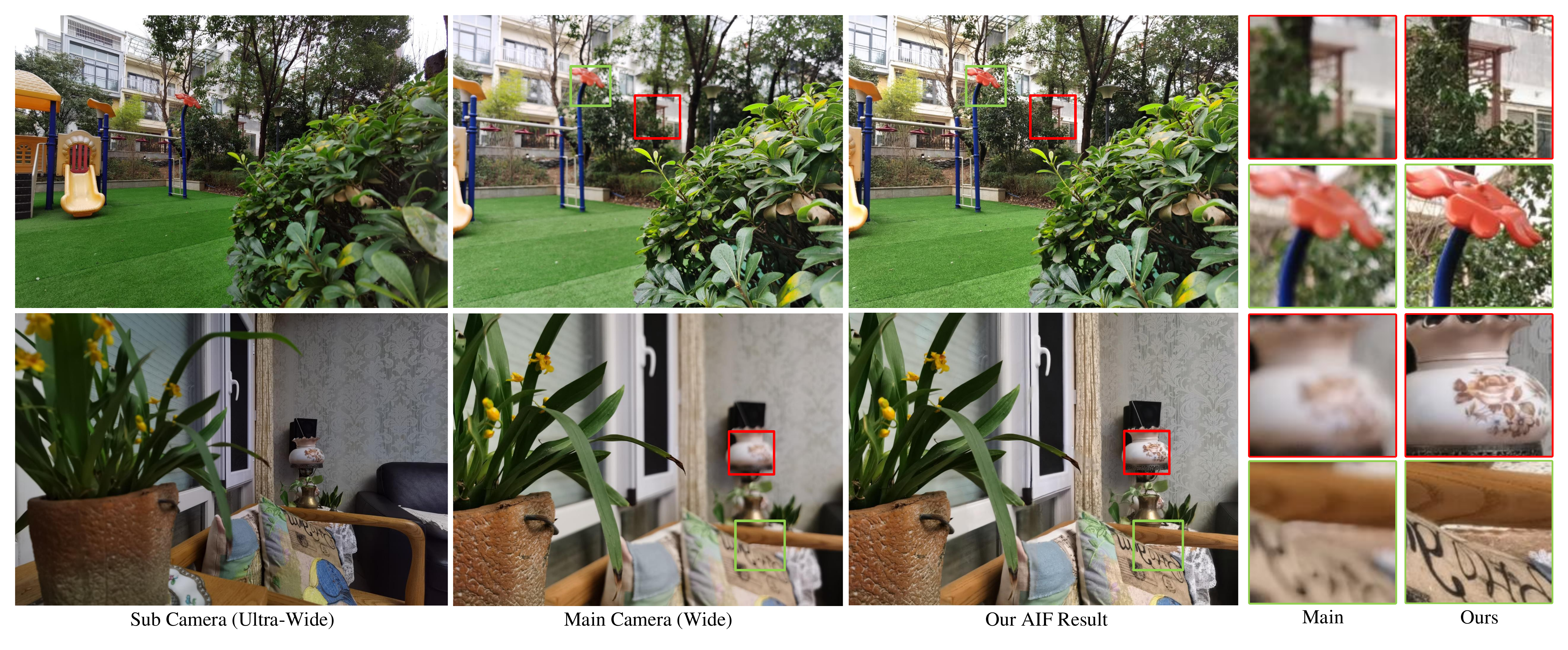}

        \caption{\textbf{All-in-Focus (AIF) photo synthesis from a smartphone camera pair}. We present a novel approach \ourmethod that operates two images captured simultaneously and respectively by a wide camera and an ultra-wide camera to repair large defocus blur (in the main lens) with clear contents (from the ultra-wide lens), enabling AIF synthesis with one click on the phone. 
        }
        \label{fig:teaser}
    \end{figure*}

	\section{Introduction}\label{sec:intro}
	\IEEEPARstart{A}{ll-in-Focus} (AIF) synthesis is commonly used in photography to keep everything sharp in a scene. To achieve AIF using a regular lens, one can decrease either the aperture size or the focal length. However, in smartphone photography, the focal length or the aperture size of the camera is fixed. Therefore, when the distance between the main camera and the foreground object is close, the lens will inevitably produce shallow depth-of-field (DOF) where either the foreground or background region is out-of-focus.
	
    In AIF synthesis, current approaches fuse the focal stack~\cite{agarwala2004interactive,pertuz2012generation,lee2016robust}, \textit{i.e.}, a set of images shot by the same camera at different focal distances, to generate an AIF photo. However, capturing a focal stack is time-consuming and requires repetitive manual refocusing. Can AIF synthesis be made simpler? We show that an additional camera can largely simplify AIF synthesis and extend the limited DOF of the main camera. 
	
\begin{figure}
            \centering
            \includegraphics[width=0.467\textwidth]{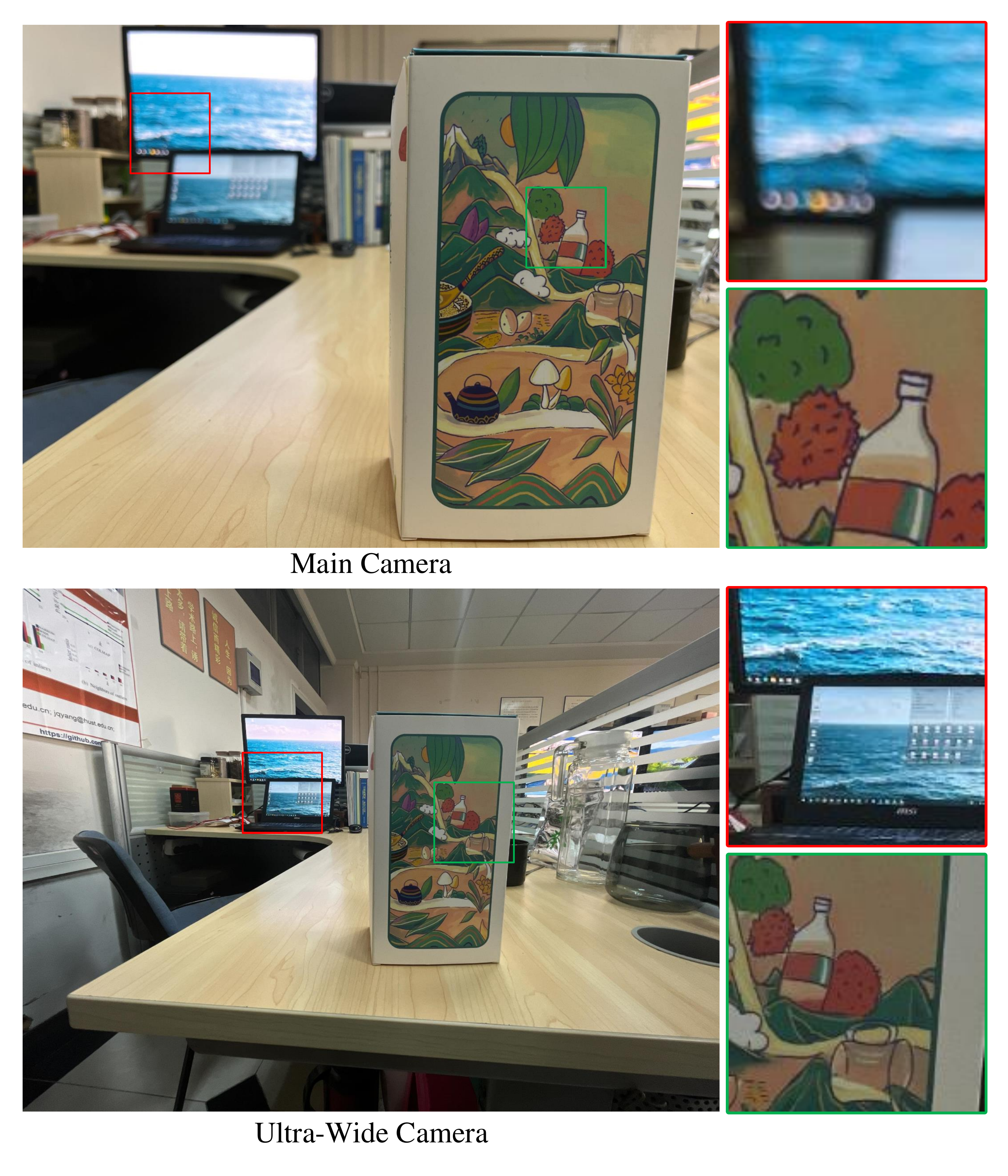}
            
        \caption{{\textbf{Features of the smartphone main/ultra-wide camera pair}. The main lens can capture a scene with good quality, but it has a shallow depth-of-field (the red square), which results in defocus blur. Therefore, we collect the image captured by the ultra-wide lens, which provides wider depth-of-field images (the green square) compared with the main lens. 
        }}
        \label{fig:lens}
    \end{figure}

	Specifically, telephoto/wide lens or wide/ultra-wide lens pair on smartphones constitutes the regular main/sub camera combination. Here we focus on the setting of wide/ultra-wide lens. As shown in  Fig.~\ref{fig:lens}, the wide-angle main lens produces high-quality details with harmonious colors, but it has a rather shallow DOF. An ultra-wide-angle lens has a small aperture and a short focal length, which results in a large DOF and an almost sharp image. {It is thus natural to seek whether the ultra-wide-angle lens can help recover missing details in the main lens to achieve AIF synthesis.} The advantages of wide/ultra-wide cameras in AIF synthesis are: 1) two cameras can work simultaneously, which is faster than adjusting focal distances with a single camera; 2) the second camera provides additional information for restoring defocused regions. {When we capture the main/ultra-wide image pair, we only need to adjust the focal distance of the main lens and leave the ultra-wide lens to the smartphone defaults. The hardware characteristics of the main/ultra-wide pair are used to synthesize an all-in-focus image with good quality.}
	
\begin{figure}
      \centering
      \includegraphics[width=\linewidth]{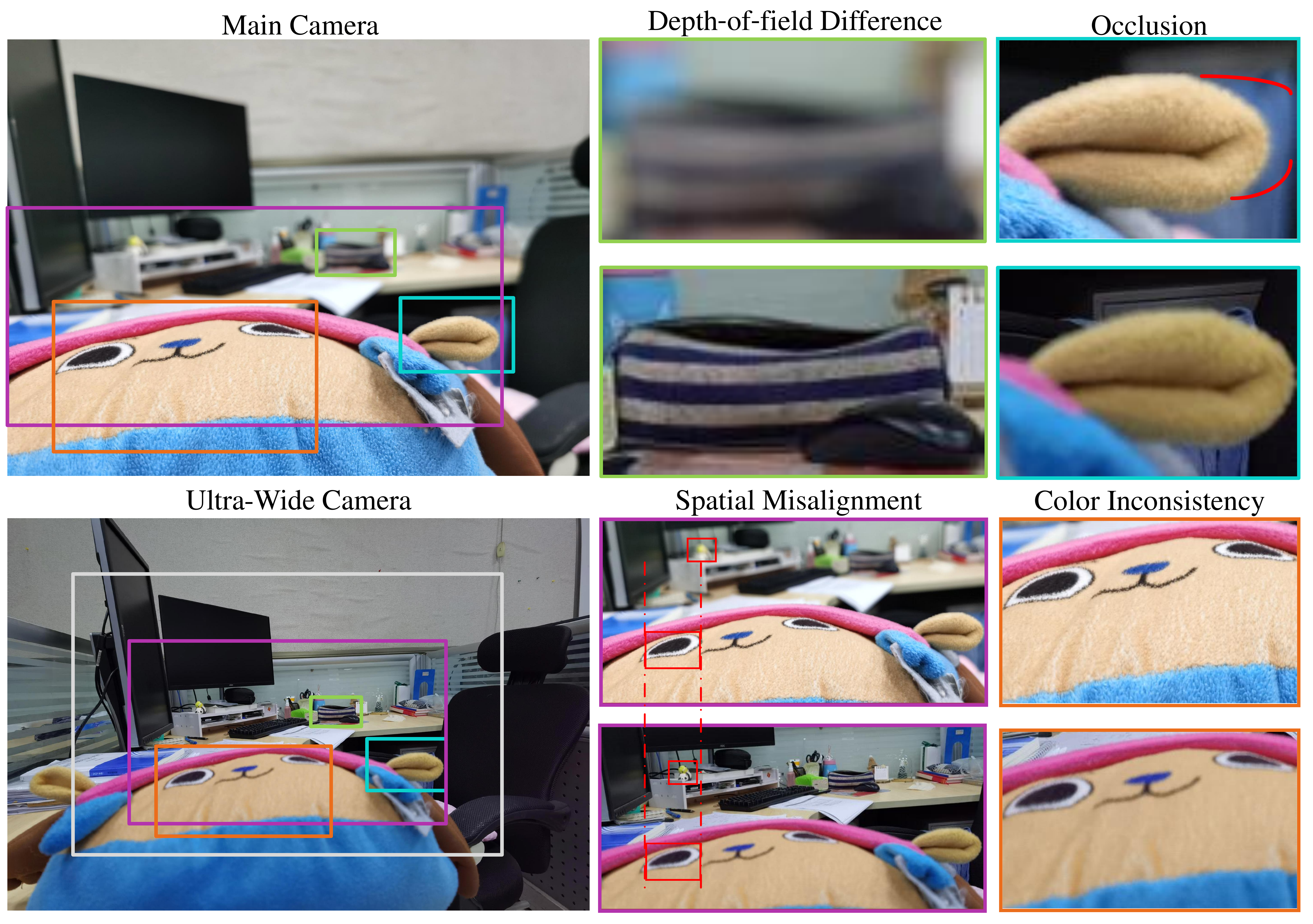}
      \caption{\textbf{Visual challenges of AIF synthesis 
      using the main/ultra-wide image pair from a smartphone}. The wide and ultra-wide images
      exhibit significant visual differences in the field of view (FoV) and resolution. 
      Although the main (wide) and ultra-wide images capture  
      overlapped scenes (the white box),
      they differ in multiple aspects, including the occlusion (the red-curve area in the blue box), the spatial misalignment
      (the red boxes in the purple box), the difference in the depth-of-field (the green box), and color inconsistency (the orange box).
      }
      \label{fig2}
    \end{figure}

	In this work, our goal is to extend the DOF of the main camera by exploiting the ultra-wide camera. However, as shown in Fig.~\ref{fig2}, the following visual challenges need to be solved: 1) the large spatial displacement between two photos; 2) the illumination and color difference between two cameras; 3) the high-resolution input size for inference.
	
	To process two inputs from different cameras, reference-based super-resolution methods~\cite{zheng2018crossnet,shim2020robust,lu2021masa,wang2021dual} are proposed to tackle spatial misalignment. However, they do not focus on shallow DOF scenes, which means they are not designed to deal with defocus blur. Furthermore, in dual-camera super-resolution, the distances between the lens and the objects of a scene are sufficiently large to define the objects as if they are from the same depth plane. On the other hand, our task deals with large foreground occlusions and defocus blur so that we cannot directly use dual-camera super-resolution models. Defocus deblurring methods~\cite{abuolaim2020defocus,lee2021iterative,son2021single,zamir2021restormer} aim to restore details from out-of-focus regions, but the lack of reference results in failure when the blurring amount is strong. Although current methods introduce dual-pixel image pairs, the image pairs are equivalent to stereo image pairs with a small baseline, which still provides no sharp reference for deblurring.
	
	Contrary to existing methods, we consider all problems above and present \ourmethod, a feasible framework to synthesize AIF photos, featured by spatial alignment, color adjustment, and occlusion-aware synthesis. To align spatial differences, we first apply homography-warping and flow warping. To further fix foreground occlusions, we design an occlusion-aware network with a deformable wavelet-based module aiming to generate sharp results in occluded regions.

    In addition to spatial displacement, we also address the color differences engendered by different lenses. To match the color of the ultra-wide photo to that of the main-camera photo, we propose a wavelet-based dynamic convolution network where dynamic convolution is used to predict weights conditioned on the blurred main-camera image. To better preserve information during downsampling and improve efficiency during inference, we apply wavelet transformation, which loses less information than conventional downsampling. Finally, the occlusion-aware network fuses sharp regions from the aligned images and outputs AIF results.

    To train our network, we collect a dataset of $2686$ scenes. We capture three photos on each scene: a background-blurred main-camera photo focused on foreground, a foreground-blurred main-camera photo focused on background, and a sharp ultra-wide sub-camera photo captured with a small aperture and a short focal length. Since the focal length and the aperture of the smartphone camera are fixed, we cannot obtain AIF ground truths for the main image directly. Therefore we resort to a fusion-based approach to synthesize a reference-based smartphone AIF dataset. In particular, multi-focus image fusion~\cite{qiu2019guided} is used to produce sharp ground truths by fusing the two main-camera photos.

    As shown in Fig.~\ref{fig:teaser}, we are the first to explore the multi-camera module in smartphone AIF synthesis. The off-the-shelf ultra-wide camera provides the defocused main photo with sharp guidance. To overcome the alignment issue, we propose a viable framework to alleviate blurring artifacts. We conduct experiments to compare our solution with existing dual-camera super-resolution and defocus deblurring approaches. We observe that current baselines are not good enough for AIF synthesis. Our framework \ourmethod is tailored to AIF synthesis and outperforms other approaches qualitatively and quantitatively. It can serve as a strong baseline for AIF synthesis from main/ultra-wide camera pair.
    
	Our main contributions include the following:
	\squishlist
	\item To our knowledge, we are the first to introduce the task of AIF synthesis from the main and ultra-wide camera pair;
	\item A strong baseline that leverages an occlusion-aware framework which fuses the main image, the spatial-color aligned ultra-wide image, and the refined main image to produce a pleasant AIF result;
	\item We collect an AIF dataset with quadruplet samples where each is composed by a pair of main-camera images that respectively focus on foreground and background, an ultra-wide-camera image, and a synthetic AIF image used as ground truth.
	\squishend
	
	\section{Related Work}\label{relwork}
    \subsection{All-In-Focus Synthesis}
    Current AIF synthesis approaches use a focal stack to composite the AIF image~\cite{agarwala2004interactive,pertuz2012generation,lee2016robust}. Typically multi-focus fusion consists of focus measure and image fusion. For sharpness measure, one can apply spatial cues~\cite{chen2017robust}, multi-scale decomposition~\cite{gangapure2015steerable}, and sparse representation~\cite{zhang2018robust}. Then fusion methods such as adaptive noise-robust fusion~\cite{pertuz2012generation} are used to selectively blend the focal stack. Apart from common solutions, the AIF image can also be reconstructed by light field synthesis~\cite{kodama2013efficient}. 
    Since capturing focal stack requires sweeping the focal plane, depth from focus~\cite{suwajanakorn2015depth,surh2017noise,sakurikar2017composite} has been proposed to perform depth estimation along with AIF synthesis. 
    
    Recently deep learning-based image fusion is used for AIF synthesis. The focus measure and the fusion rule can be learned from a deep network~\cite{liu2017multi,xiao2020global}. Deep unsupervised learning is widely used to fuse a focal stack without explicit supervision~\cite{xu2020u2fusion,zhang2021mff}. In addition, depth-from-focus networks~\cite{maximov2020focus} adopt supervised learning to predict an AIF image and a depth map from corresponding synthetic ground truth. {A joint multi-level feature extraction-based CNN~\cite{zhao2018multi} is proposed to provide a natural enhancement. A simplified stationary wavelet transform using Harr filter~\cite{liu2019new} is proposed to achieve fast fusion speed.}
    
    Although focal stack is effective in generating AIF results,
    it is time-consuming and inefficient to 
    manually change focal planes and capture required image sequences using the same smartphone lens.
    Therefore, we propose a user-friendly AIF synthesis routine to produce an AIF image from a main and ultra-wide image pair that can be captured at the same time. {Traditionally the focal stack or the light field images are shot by the same camera, so the captured images do not suffer from misalignment in space and color. Compared with the common multi-focus fusion methods~\cite{zhao2018multi,liu2019new}, our inputs are two misaligned images from two different cameras instead of the common defocused images from the same camera.}

    \subsection{Defocus Deblurring}

    In addition to direct AIF synthesis by image fusion, defocus deblurring is also feasible to restore a sharp image from a bokeh image~\cite{luo2020bokeh,Peng_2022_CVPR}. Traditional methods apply a two-stage technique~\cite{levin2007image,park2017unified,krishnan2009fast,lee2019deep}, where defocus estimation~\cite{park2017unified,lee2019deep,kumar2019simultaneous,shi2015just,liu2020estimating} is first executed on a pre-defined blur model, then the predicted defocus map helps to deblur the image by non-blind deconvolution~\cite{levin2007image,krishnan2009fast}.
    
    Instead of using deep networks to predict a defocus map, recent deep learning methods directly deblur the image~\cite{abuolaim2020defocus,lee2021iterative,son2021single,zamir2021restormer}. Therefore, a defocus deblurring dataset~\cite{abuolaim2020defocus} consisting of dual-pixel images is introduced for training. Better network designs have been proposed, such as iterative adaptive~\cite{lee2021iterative} and kernel-sharing parallel atrous~\cite{son2021single} convolutions. With the help of dual-pixel images, multi-task learning methods such as defocus estimation~\cite{xin2021defocus} and predicting dual-pixel views~\cite{abuolaim2022improving} are also proved to benefit defocus deblurring. Synthetic datasets such as dual-pixel video sequences~\cite{abuolaim2021learning} and depth images~\cite{pan2021dual} have also been used as assistance for defocus deblurring. The transformer architecture is also applied to defocus deblurring~\cite{zamir2021restormer} and achieves the state of the art. {In addition, light field data~\cite{ruan2021aifnet} can be applied to synthesize defocus blur from a set of sharp images, and a network is proposed to deblur a single spatially varying defocused image.}
    
    Compared with reference-based AIF synthesis, defocus deblurring also restores a sharp image. However, it only utilizes a single image or a dual-pixel image pair instead of a sharp reference image, which is not sufficient to recover details from large blur. {Compared with the restoration method from light field~\cite{ruan2021aifnet}, our method requires two complimentary inputs, while the light field dataset only enables single image deblurring by synthesize one defocused image from mutiple light field images. Furthermore, the light field dataset is captured by the same camera, so the light field images are aligned, which is different from our AIF dataset.} 
    
    \subsection{Dual Camera Applications}
    Dual camera has been used in various low-level applications such as reference-based super-resolution~\cite{zheng2018crossnet,shim2020robust,lu2021masa,wang2021dual}, reference-guided image inpainting~\cite{zhou2021transfill}, and bokeh rendering~\cite{luo2020wavelet}. 
    Reference-based super-resolution methods align two images from different viewpoints~\cite{zheng2018crossnet,shim2020robust,lu2021masa}. Particularly, dual-camera super-resolution~\cite{wang2021dual} super-resolves the wide image with the help of the telephoto lens, which can be applied to smartphone images. 
    
    Compared with our wide/ultra-wide AIF synthesis task, dual-camera super-resolution methods are designed for two misaligned input images. However, they do not consider defocus blur or foreground occlusions. 
    In addition, reference-guided inpainting requires a mask for inputs, and bokeh rendering does not belong to image restoration task, so there is little similarity between our task and existing ones.
	
    \section{\ourmethod for All-In-Focus Synthesis}
\begin{figure}
      \centering
      \includegraphics[width=\linewidth]{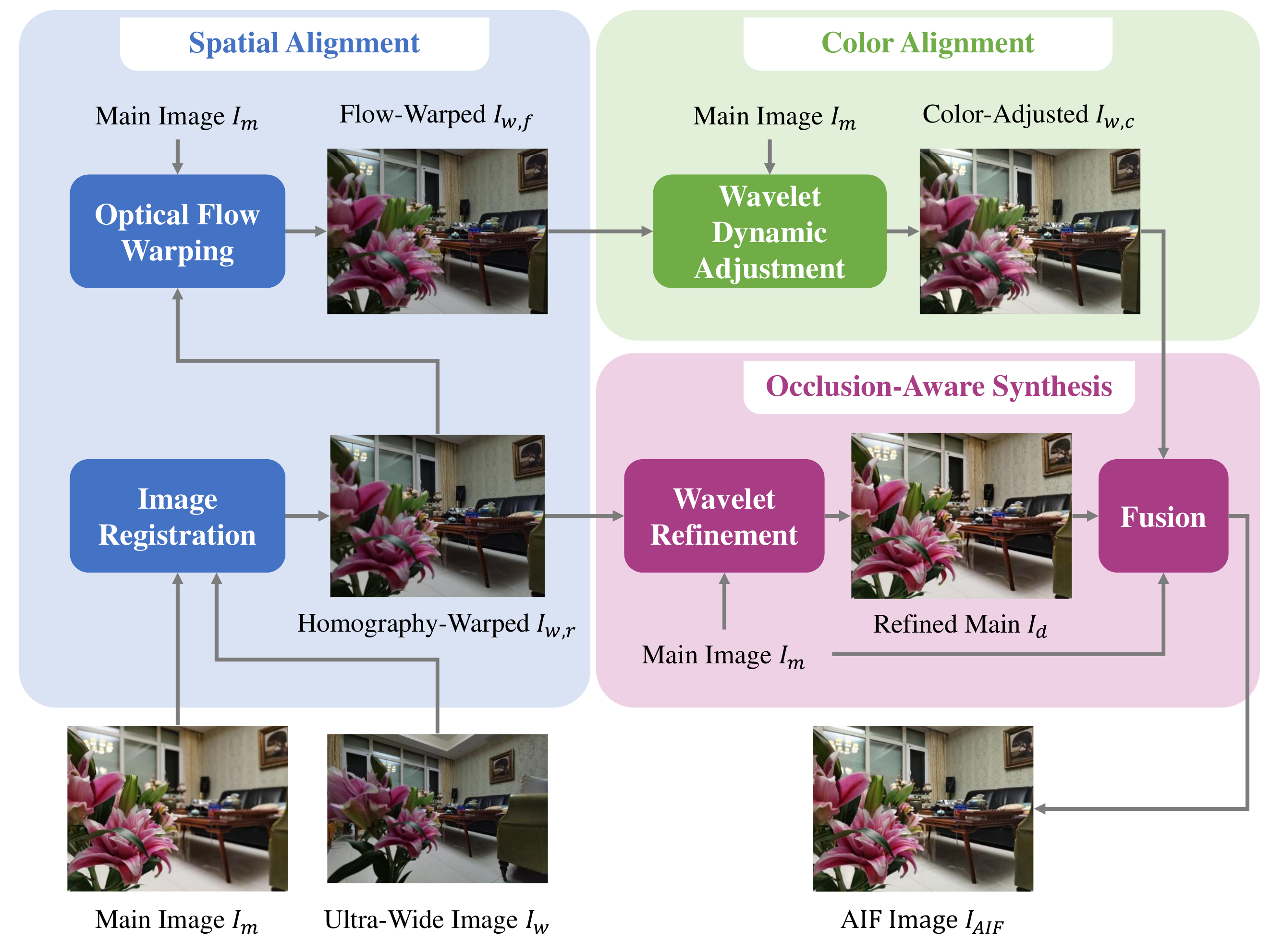}
      \caption{
      \textbf{Our framework \ourmethod includes three modules: spatial alignment, color alignment, and occlusion-aware synthesis}. 
      We use homography and flow warping to align the main/ultra-wide image pair $I_m$ and $I_w$ at the image level and the pixel level. Then we adjust the color of the warped ultra-wide image $I_{w,f}$ with a wavelet-based dynamic network. To tackle the occlusions where spatial warping fails to solve, we propose an occlusion-aware synthesis network to refine the occluded and blurred areas. The network fuses the refined image $I_d$, the main image $I_m$, and the color-adjusted ultra-wide $I_{w,c}$ to generate the AIF result $I_{AIF}$.}
      \label{fig3}
    \end{figure}

    As shown in Fig.~\ref{fig3}, our framework \ourmethod consists of three components: spatial alignment, color alignment, and occlusion-aware synthesis. To solve the spatial misalignment between the main camera $I_m$ and the ultra-wide camera $I_w$, we apply both homography warping and flow warping to output the aligned ultra-wide $I_{w,f}$. Homography warping focuses on a global scale and produces a coarse result, and flow warping is stable under parallax issues brought by different depth planes. 
    Once spatial warping is done, we also align the two images $I_m$ and $I_{w,f}$ photometrically. To
    process images of high resolution, we propose a wavelet-based network to preserve information during downsampling. Furthermore, we apply dynamic convolution to generate the color-aligned $I_{w,c}$ utilizing the reference blurred main image $I_m$. Since the dual-camera setting leads to occlusions around the edge of foreground objects, we design an occlusion-aware synthesis network, where we simultaneously find the occlusions and refine the blurred occluded area to produce $I_d$. Finally the fusion network fuses $I_m$, $I_{w,c}$, and $I_d$ to synthesize the AIF image $I_{AIF}$.
	
	\subsection{Spatial Alignment}
    We implement spatial alignment by means of homography warping and flow warping.
    Homography warping aligns the input 
    at the image level, which 
    can restore image-level attributes such as the FOV. However, in our smartphone AIF synthesis task, prominent foreground/background relationships result in a larger disparity 
    on background regions than on foreground ones. Therefore it is impossible to align the main/ultra-wide image pair $I_m$ and $I_w$ using only a single homography warping. 
    To align objects from different depth planes, we apply pixel-wise warping with an optical flow field, because we need diverse offsets to fit the increasing parallax from foreground to background. 
    
\begin{figure}
      \centering
      \setlength{\abovecaptionskip}{3pt}
        \setlength{\belowcaptionskip}{0pt}
      \includegraphics[width=\linewidth]{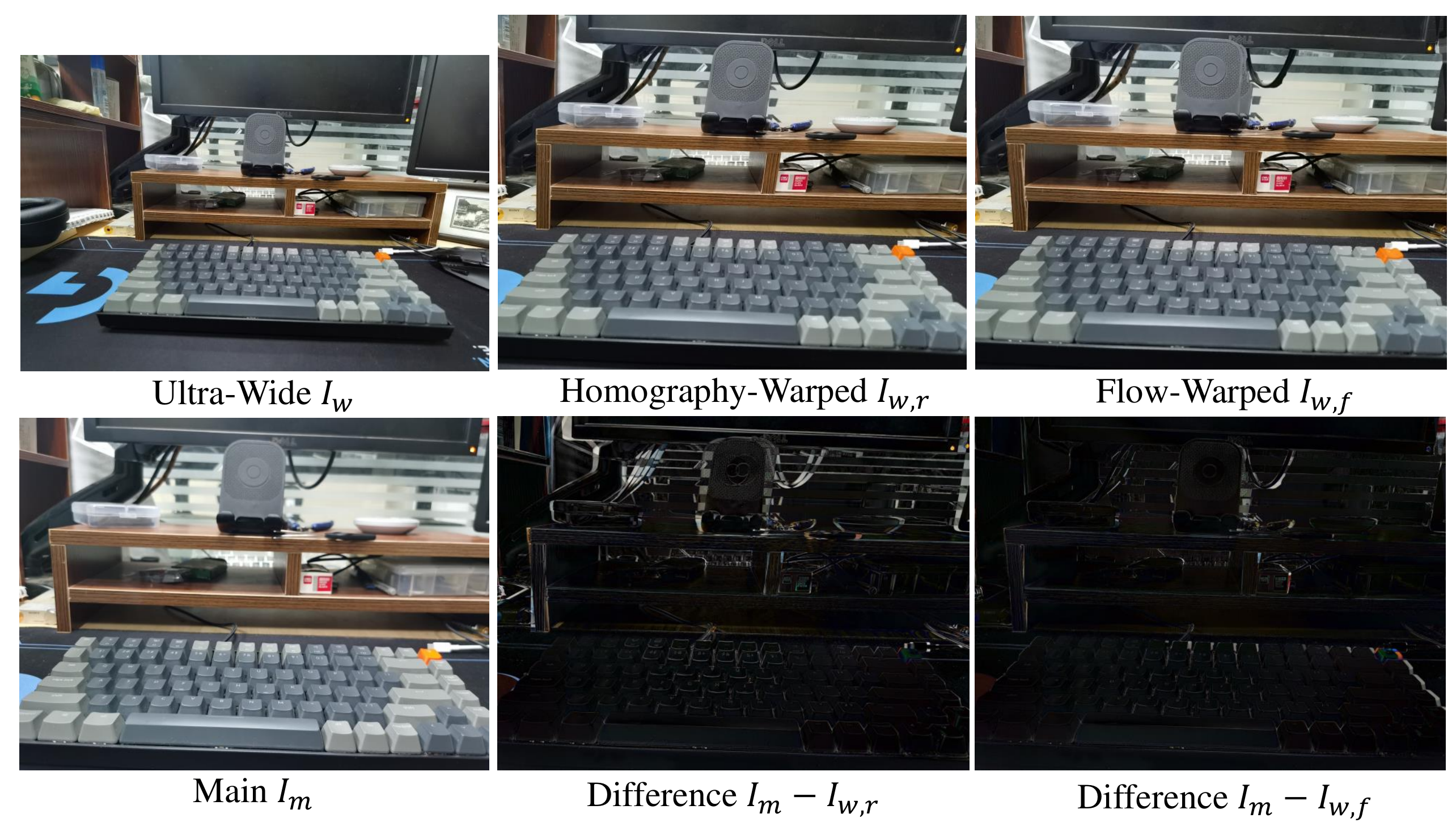}
      \caption{\textbf{The difference between a warped ultra-wide image and a main image}. $I_w$ has a different aspect ratio from $I_m$.}
      \label{spatial}
    \end{figure}

    Homography warping is executed by image registration. Image registration consists of 1) keypoints detection, 2) feature matching, 3) outlier pre-filtering, and 4) pose estimation. We locate keypoints from SIFT~\cite{lowe1999object}, and extract HardNet~\cite{mishchuk2017working} descriptors to compute an initial correspondences set between $I_w$ and $I_m$. Then we apply a pre-trained outlier rejection network NM-Net~\cite{zhao2019nm} to filter unreliable correspondences in the initial set.
    Finally, $I_w$ is warped by a single homography matrix estimated from the filtered correspondences using RANSAC~\cite{fischler1981random}.
    Homography warping provides a coarse image-level alignment for our pipeline. 
    The warping results $I_{w,r}$ and $I_m$ still have unaligned regions due to large parallax. 
    Therefore, we apply a pixel-wise warping alignment with an optical flow field.
    Pixel-wise warping is insensitive to varying parallax from different depth planes. 
    We apply the robust RAFT~\cite{teed2020raft} to estimate the flow map $Y_{I_m \rightarrow I_{w,r}}$ for aligning $I_{w,r}$ to $I_m$. 
    As shown in Fig.~\ref{spatial}, homography alignment provides a coarse image-level result, which is effective in dealing with FOV discrepancy, and the flow warping exhibits more adaptability in aligning objects from different depth planes such as the speaker in $I_{w,r}$.
    
	\subsection{Wavelet-Based Dynamic Color Alignment}
    In image alignment, we not only consider spatial differences, but also color discrepancies. Since two different cameras are involved,
    color alignment must be considered. Specifically, smartphone AIF synthesis defines a special scenario for color alignment. In our pipeline we have a reference blurred input $I_m$ as guidance, which is different from the previous single image enhancement~\cite{gharbi2017deep}. Furthermore, the to-be-adjusted $I_{w,f}$ is spatially misaligned but only in occlusions, so it is not the case of style transfer~\cite{gatys2016image}.
    \begin{figure}
      \centering
      \includegraphics[width=\linewidth]{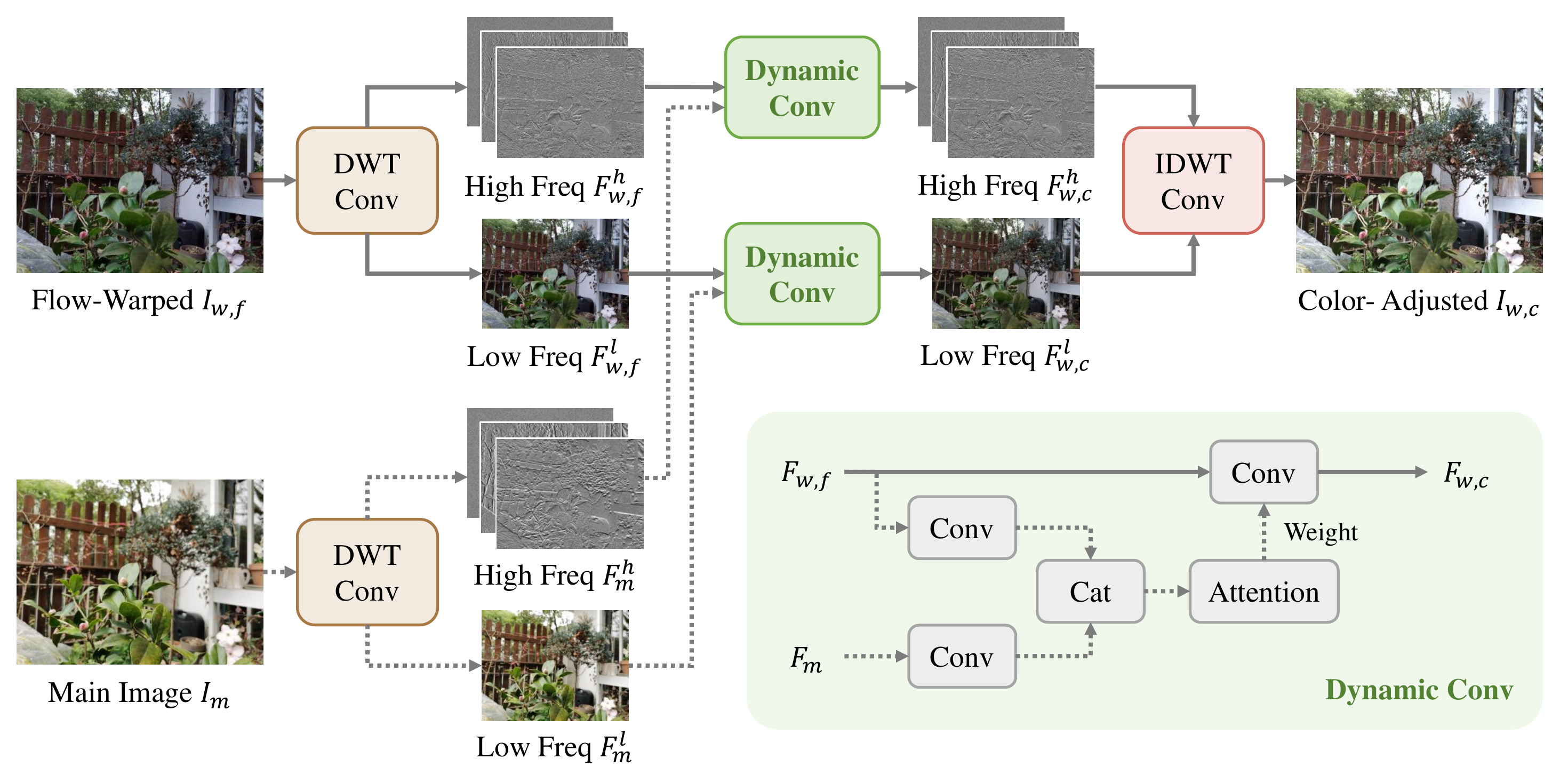}
      \caption{\textbf{WDC-Net for color alignment}. We use DWT to downsample the RGB image into high and low frequency contents $F^h$ and $F^l$. To utilize the reference image, we apply dynamic convolution on high and low frequency to adjust color transform parameters based on the reference $I_m$. The adjusted low frequency component is reconstructed to the original resolution $I_{w,c}$ guided by high frequency details.}
      \label{color}
    \end{figure}
    
    To use the reference main camera $I_m$, we propose a wavelet-based dynamic color alignment network WDC-Net. Particularly, we integrate dynamic convolution~\cite{chen2020dynamic} with discrete wavelet transformation (DWT) and inverse DWT. As shown in Fig.~\ref{color}, DWT decomposes the RGB image into high and low frequency contents, and the resolution of output is half of the input size. It has been revealed that in image-to-image translation, the color transformation relies more on the low frequency contents~\cite{liang2021high}. Therefore, we use DWT to align the color in low frequency and refine the low frequency result with high frequency cues. Furthermore, we expect that the main image can 
    guide color alignment, so we apply dynamic convolution such that the color-transform-convolution weight is flexible based on the input main image.
	
    \subsection{Occlusion-Aware Synthesis}
    Now that we discuss the spatial alignment performed by homography and flow warping, we still have the occlusion problem. Occlusion arisen from two perspectives is not rare. However, it is worth emphasizing in smartphone AIF synthesis. 
    As shown in Fig.~\ref{fig2}, 
    when the camera is close to the foreground object, the parallaxes in each depth plane are significantly different, which brings more occlusions
    than other tasks such as dual-camera super-resolution. 
    
\begin{figure*}[h]
      \centering
      \setlength{\abovecaptionskip}{3pt}
        \setlength{\belowcaptionskip}{0pt}
      \includegraphics[width=\linewidth]{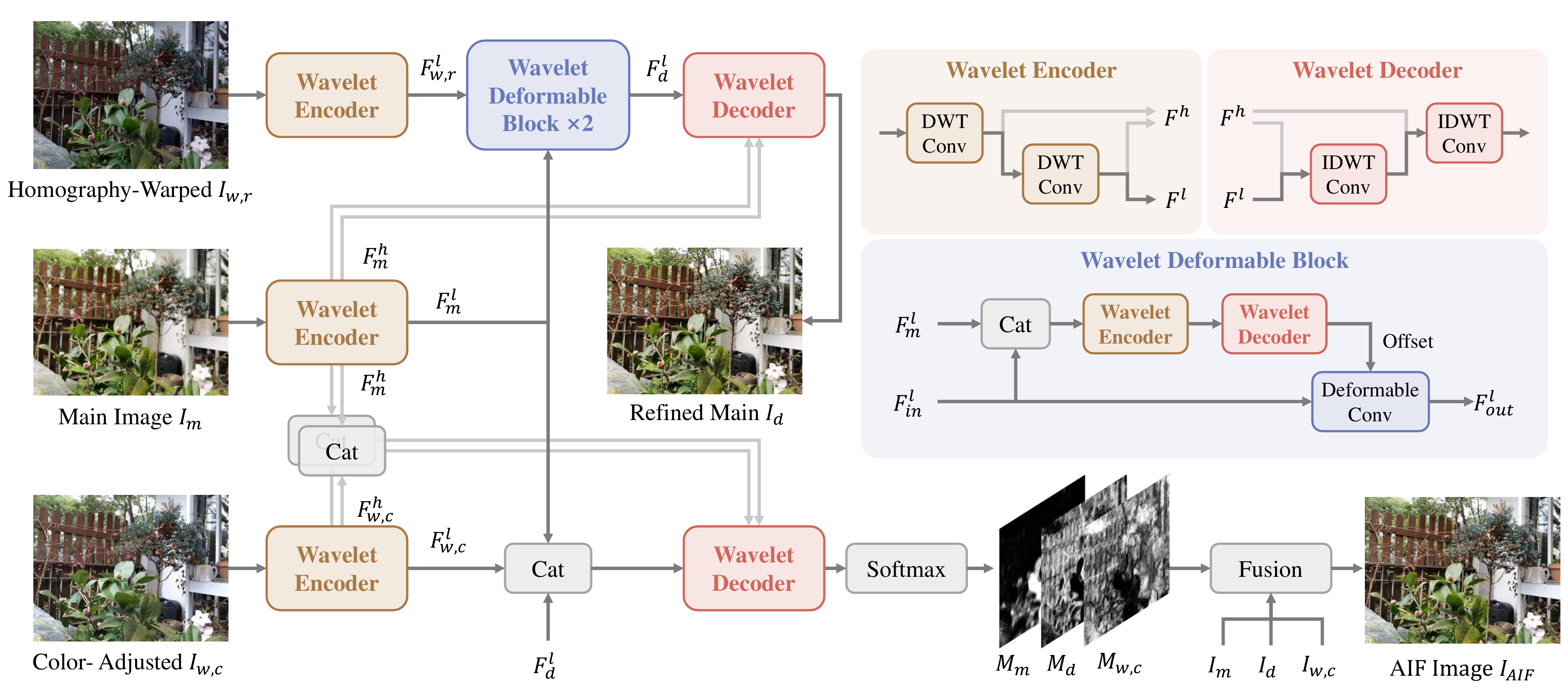}
      \caption{\textbf{Occlusion-aware synthesis network OAS-Net}. We modify the common encoder-decoder with wavelet transform, and we integrate the wavelet encoder and decoder into deformable convolution, where we predict the offset in the frequency domain. The three wavelet encoders of the three inputs ($I_{w,r}$, $I_m$, and $I_{w,c}$) share the same parameters. The two decoder branches output the fusion masks ($M_m$, $M_d$ and $M_{w,c}$) for AIF synthesis and the refined main image $I_d$ for occluded regions $M_d$. }
      \label{network}
    \end{figure*}

    Taking occlusions into account, when the main camera focuses on foreground objects, the ultra-wide image cannot obtain corresponding sharp results in the occluded regions. 
    The flow warping in spatial alignment tends to produce 
    poor flows in occluded areas, so we first predict a confidence map $M_c$ by forward-backward consistency check~\cite{chen2016full} as 
    \begin{equation}
        \begin{aligned}
         M_c = \left \| Y_{I_{w,r} \rightarrow I_m}(p)+Y_{I_m \rightarrow I_{w,r}}(p+Y_{I_{w,r} \rightarrow I_m}(p)) \right \|_2\leq 1\,,
        \end{aligned}
        \label{consistency}
    \end{equation}
    where $p$ is considered as a point in $I_{w,r}$, $Y_{I_{w,r} \rightarrow I_m}$
    is the flow from the ultra-wide image to the main image, 
    and $Y_{I_m \rightarrow I_{w,r}}$ is the flow from the main image to the ultra-wide image. The intuition
    behind
    the consistency check is that, if an estimated flow is correct, then when a corresponding point is warped twice by the forward flow and backward flow, the output should be close to zero.
    To synthesize an AIF output, we not only need to find where occlusions occur but also to acquire clear results in those areas.
    We address them with an occlusion-aware synthesis network OAS-Net, as shown in Fig.~\ref{network}.

    We use an encoder-decoder structure with an efficient twist that replaces downsampling and upsampling operations with DWT and inverse DWT. DWT decomposes the feature map into high and low frequency contents, and IDWT reintegrates the frequency back into the feature map, which is similar to the DWT and IDWT in Fig.~\ref{color}. Since we predict the fusion masks for AIF synthesis, the fusion mask is flat in most parts, resulting in less information in high frequency contents. Therefore the network can predict the fusion masks mainly in low frequency, then restore the masks to the original resolution with high frequency cues. In addition, the input resolution of a smartphone camera is high, we utilize wavelet transformation so that the mask can be predicted in lower resolution with less information loss than normal pooling layers.
    
    As we predict the occluded areas, we simultaneously refine the corresponding blurred background with the help of homography-warped ultra-wide image $I_{w,r}$. We apply deformable convolution~\cite{dai2017deformable} to align the features of $I_{w,r}$ and $I_m$, then we reconstruct the refined image $I_d$ based on the aligned features. Compared with previous works~\cite{shim2020robust,tian2020tdan}, we predict deformable weights with the direct use of DWTs and inverse DWTs. We replace convolution with wavelet decomposition in deformable weights predictions because DWT is helpful in capturing global context and in preserving more information in downsampling than
    standard convolution or pooling. With the refined image $I_d$, the main image $I_m$, and the color-aligned ultra-wide $I_{w,c}$, we fuse them with their corresponding fusion masks $M_d$, $M_m$, and $M_{w,c}$ as
    \begin{equation}
        \begin{aligned}
         I_{AIF} = I_d\cdot M_d+I_m\cdot M_m+I_{w,c}\cdot M_{w,c}\,,
        \end{aligned}
    \end{equation}
    where $I_{AIF}$ is the final AIF result. 
    {We apply a softmax function to limit the network training, so $M_{w,c}+M_d+M_m=1$.}
    $M_d$ is supervised by $M_c$, because $M_d$ mostly represents occlusions and low-confidence flow warping results usually occur in occluded regions.

    % \squishend

	\subsection{Model Training}
    \label{sec:model_training}

    \noindent\textbf{Loss Functions.}
    For WDC-Net, we use the following loss:
    \begin{equation}
        \begin{aligned}
        \mathcal{L}_{\mathit{WDC}} &= \mathcal{L}_{1}(I_{w,c}\cdot M_c,I_{gt}\cdot M_c)\\
        &+\mathcal{L}_{ssim}(I_{w,c}\cdot M_c,I_{gt}\cdot M_c)\,,
        \end{aligned}
    \end{equation}
    where $L_{1}$ is the $\ell_1$ loss, and $L_{ssim}$ is the structural similarity (SSIM) loss~\cite{DBLP:journals/tci/ZhaoGFK17}. $I_{gt}$ is the AIF ground truth image, and $M_c$ is the confidence mask of optical flow computed from forward-backward consistency check. We use $M_c$ to mitigate the influence of wrong warped regions.
    
    For OAS-Net, 
    we have the loss function
    \begin{equation}
        \begin{aligned}
        \mathcal{L}_{\mathit{OAS}} &= \mathcal{L}_{1}(I_{AIF},I_{gt})+\mathcal{L}_{ssim}(I_{AIF},I_{gt})\\
        &+\lambda\cdot\mathcal{L}_{1}(I_{d},I_{gt})+\lambda\cdot\mathcal{L}_{ssim}(I_{d},I_{gt})\\
        &+\delta\cdot\mathcal{L}_{bce}(M_{m},M_{fuse}) \\
        &+\delta\cdot\mathcal{L}_{bce}(M_{d},(1-M_{fuse})\cdot M_c)\\
        &+\delta\cdot\mathcal{L}_{bce}(M_{w},(1-M_{fuse})\cdot (1-M_c))\,,
        \end{aligned}
    \end{equation}
    where $L_{bce}$ is the binary cross entropy loss. $M_{fuse}$ is the fusion mask for generating AIF ground truth $I_{gt}$ as
    \begin{equation}
        \begin{aligned}
        I_{gt}=M_{fuse}\cdot I_{m}^{fg}+(1-M_{fuse})\cdot I_{m}^{bg}\,,
        \end{aligned}
    \end{equation}
    where $I_{m}^{fg}$ and $I_{m}^{bg}$ are the main images focused on foreground and background, respectively. We describe the collections of $I_{m}^{fg}$, $I_{m}^{bg}$, $M_{fuse}$, and $I_{gt}$ in Sec.~\ref{sec:dataset}. We use the binary cross entropy loss because we use a softmax function at the end of mask prediction, and the three mask ground truth images can all be defined as hard masks. $(1-M_{fuse})*M_c$ is the ground truth of $M_d$ because we want the refined image to fill in the low-confidence regions but leave out parts where the main image remains clear. $M_{fuse}*(M_c-1)$ is set as the ground truth of $M_w$ because we want to ensure the three ground truth images add up to $1$.
    
    \vspace{5pt}
    \noindent\textbf{Implementation Details.}
    We implement our model by PyTorch~\cite{paszke2017automatic}. $\lambda$ and $\delta$ are set to $0.5$ and $0.1$, respectively. We apply one DWT and one IDWT in WDC-Net. Two sets of DWT and IDWT are used in the encoder-decoder of OAS-Net, and three sets of DWT and IDWT are used in the dynamic offset estimator in the deformable convolution block. When training WDC-Net, we do not downsample the input beforehand. We downsample the image by a factor of 4 before training the OAS-Net. Both networks are trained for $40$ epochs using the Adam optimizer~\cite{kingma2015adam}. WDC-Net uses a batch size of $8$, and OAS-Net uses a batch size of $2$. Our experiments are conducted on a single NVIDIA GeForce GTX 1080 Ti GPU, while some of the baseline experiments require an NVIDIA GeForce GTX 3090 GPU.

    \section{Dataset Collection} 
    \label{sec:dataset}
    To train our AIF photo synthesis network, we collect the main/ultra-wide image pair for input, and the ground truth AIF image as supervision.
    We capture the smartphone AIF dataset from Huawei Mate 30 Pro, which uses an ultra-wide camera in its module. 
    \begin{figure}
      \centering
      \includegraphics[width=\linewidth]{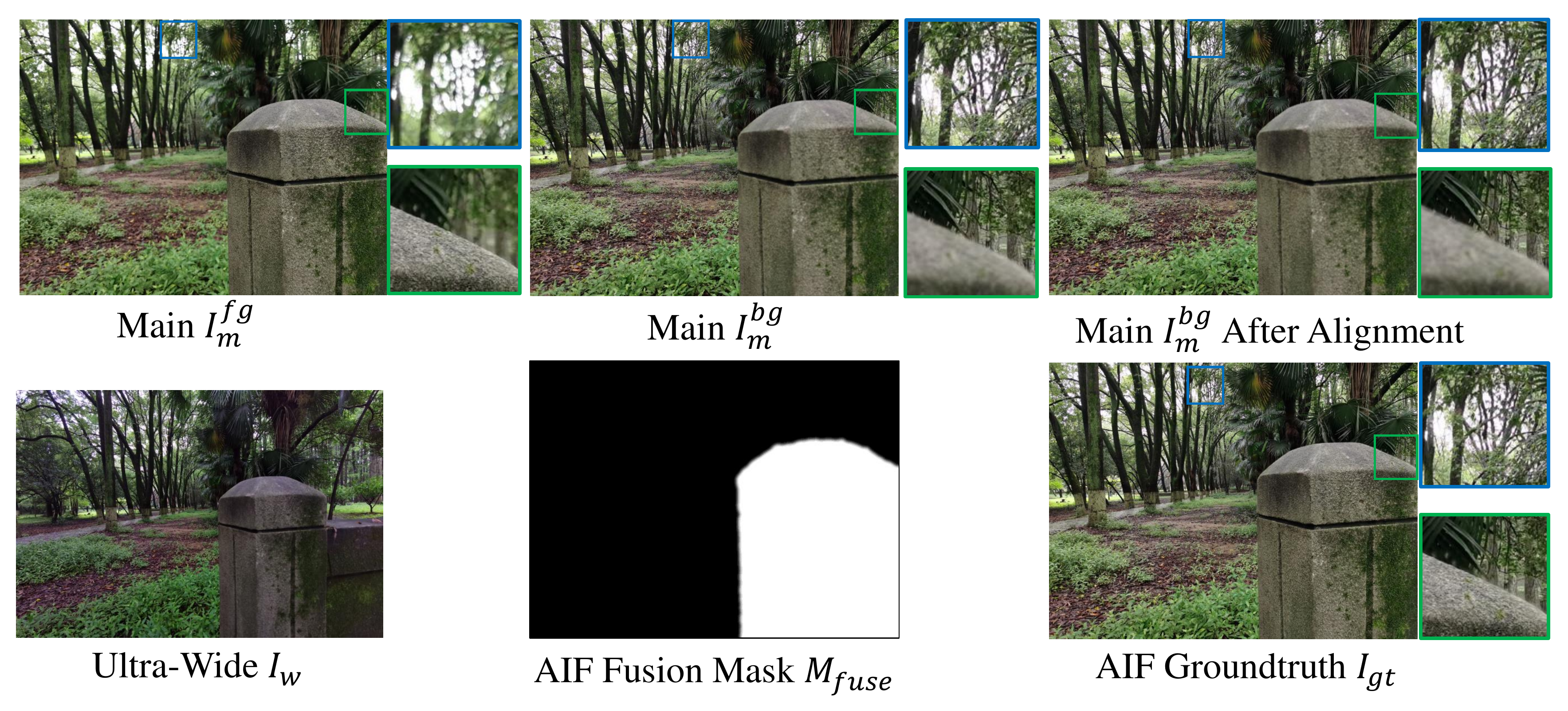}
      \caption{\textbf{An example of a set of training images}. The main images $I_{m}^{fg}$ is focused on foreground, and $I_{m}^{bg}$ is focused on background, as shown in the blue and green boxes. Since the focus breathing effect causes wider FOV in $I_{m}^{bg}$ (the green box), we align $I_{m}^{bg}$ to $I_{m}^{fg}$ using flow warping. We also capture the ultra-wide image $I_w$, and the AIF ground truth image$I_{gt}$ is generated from fusing the two main images.}
      \label{fig:dataset}
    \end{figure}
    To obtain an image set, we first use the main camera to shoot two images; one focuses on the foreground, and one on the background. Then we switch to the ultra-wide camera to capture an image $I_w$ following the default camera setting. We use a tripod to avoid camera shaking. We keep the exposure settings same for the two lenses, because we intend to simulate the situation where the two cameras work simultaneously.
    
    Now that we have three images per scene, we need to align the two main images due to focus breathing (a phenomenon that the FOV becomes narrow when focusing on a closer object). As shown in Fig.~\ref{fig:dataset}, the two main images $I_{m}^{fg}$ and $I_{m}^{bg}$ have view angle differences, where $I_{m}^{bg}$ has a wider FOV. Therefore, we apply optical flow warping to align $I_{m}^{bg}$ to $I_{m}^{fg}$.
    We take the two main camera images $I_{m}^{fg}$ and the aligned $I_{m}^{bg}$ as inputs and apply a multi-focus image fusion method~\cite{qiu2019guided} to synthesize the AIF ground truth $I_{gt}$ for training. In addition, we acquire the fusion mask $M_{fuse}$ for the two main images. 
    
    In all, we collect a dataset with $2686$ indoor and outdoor scenes, with $2686*2=5372$ image pairs.
    We split the dataset into $5000$ training image sets and $372$ evaluation image sets. Each set consists of $I_{m}^{fg}$, $I_{m}^{bg}$, $I_w$, $M_{fuse}$, and $I_{gt}$, as shown in Fig.~\ref{fig:dataset}. The ultra-wide image has a resolution of $3840*2592$, and other images have a resolution of $3648*2736$. The ultra-wide camera has a focal length of $18$ mm and an aperture size of $f/1.8$, and the main camera has a focal length of $27$ mm and an aperture size of $f/1.6$. To demonstrate the generalization of \ourmethod, we also capture $120$ image pairs from iPhone13 to further solidify our claim.

	\section{Results and Discussions}\label{sec:results}

	\subsection{Baseline Approaches}
    We choose state-of-the-art baselines that are
    closely related to our new task, consisting of two types of approaches: 1) dual-pixel defocus deblurring: IFAN~\cite{lee2021iterative} and Restormer~\cite{zamir2021restormer} and 2) dual-camera super-resolution: DCSR~\cite{wang2021dual} and MASA-SR~\cite{lu2021masa}. 
    
    \noindent\textbf{IFAN~\cite{lee2021iterative}} is a single image defocus deblurring method which applies iterative adaptive convolutions. This model is trained on a dual-pixel dataset.
    
    \noindent\textbf{Restormer~\cite{zamir2021restormer}} addresses image restoration by using Transformer. The defocus deblurring model can be  trained on a dual-pixel dataset. 
    
    \noindent\textbf{DCSR~\cite{wang2021dual}} explores the dual camera
    super-resolution with aligned attention modules.
    
    \noindent\textbf{MASA-SR~\cite{lu2021masa}} uses matching acceleration and the spatial adaptation module to achieve reference-based super-resolution.
    
    We use their pretrained models, and finetune the models on our dataset for a fair comparison.
	
	\subsection{Qualitative Results}
\begin{figure}
      \centering
      \includegraphics[width=\linewidth]{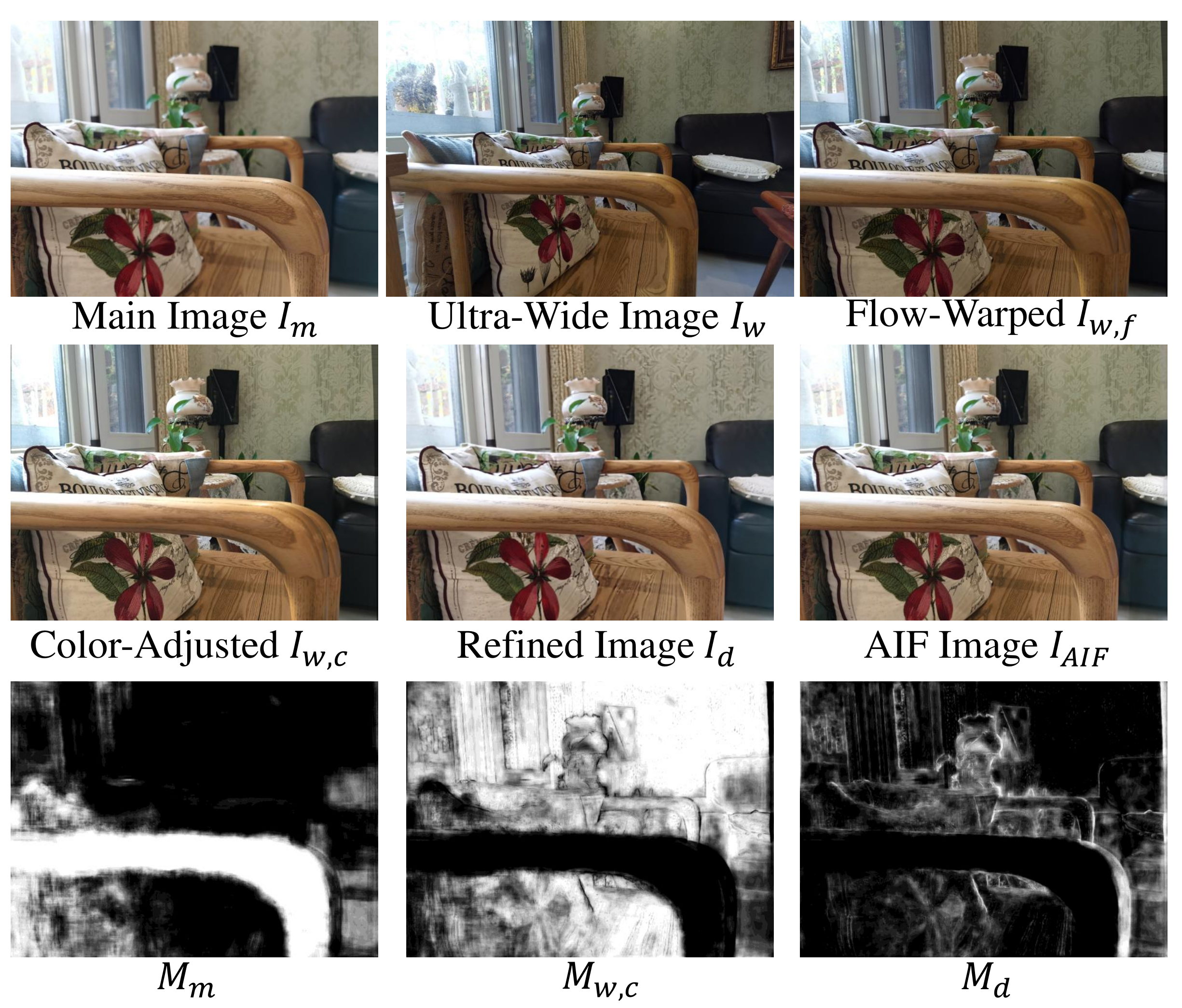}
      \caption{\textbf{Intermediate results of EasyAIF}. We demonstrate how the three components of EasyAIF work to synthesize $I_{AIF}$. The main image $I_m$ is focused on the foreground regions. $M_d$ includes the occlusions at the edges of foreground objects because the corresponding occluded regions of $I_d$ are sharper than that of foreground-focused $I_m$.}
      \label{fig:intermediate1}
    \end{figure}

\begin{figure*}
      \centering
      \setlength{\abovecaptionskip}{3pt}
      \includegraphics[width=\linewidth]{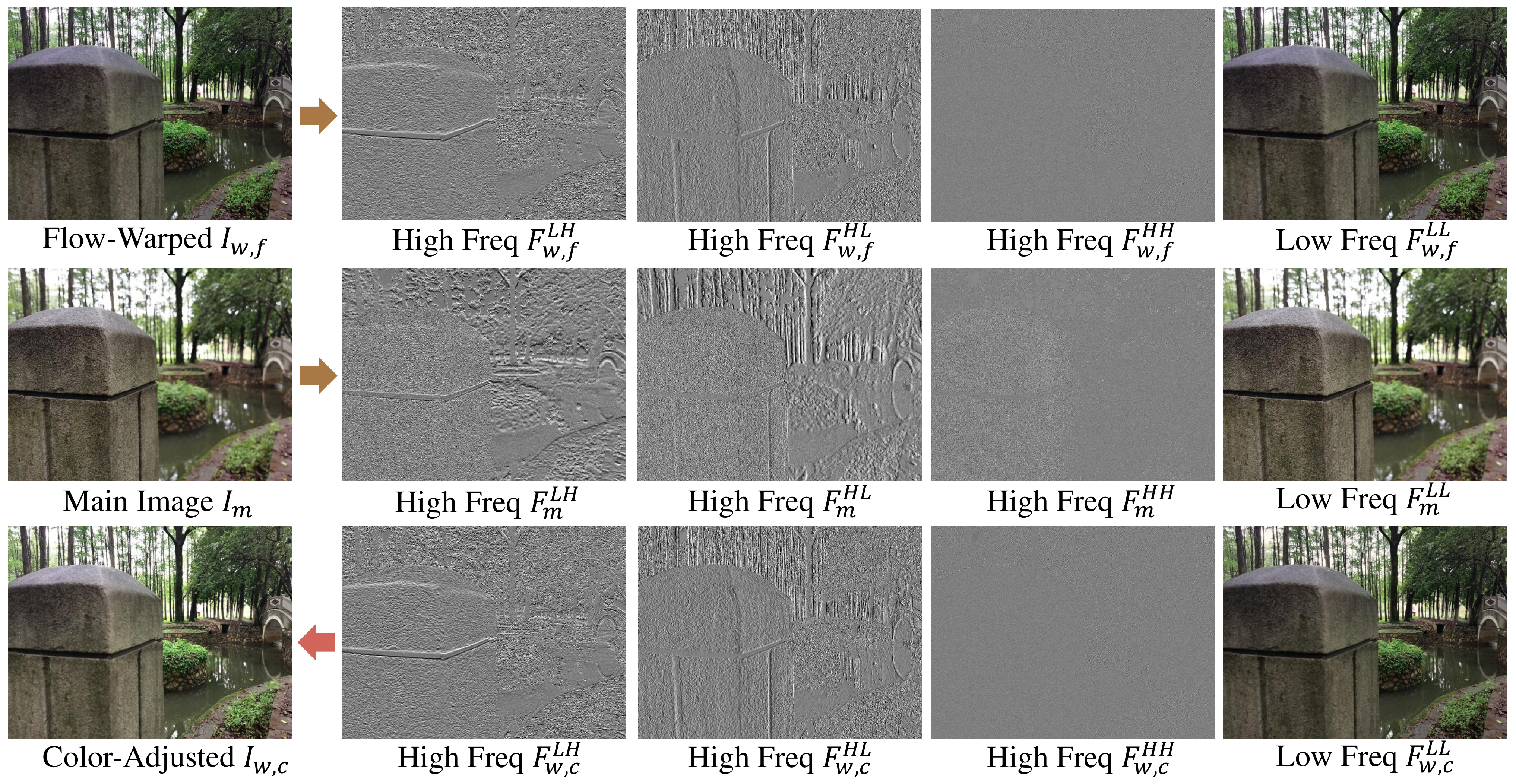}
      \caption{\textbf{Intermediate feature maps of WDC-Net}. The arrows indicate that $I_{w,f}$ and $I_m$ are the inputs, and WDC-Net outputs $I_{w,c}$ from high frequency features and low frequency features. We can observe that the high frequency feature maps store information of the edges, and the low frequency feature maps relate more to the contents.}
      \label{fig:intermediate2_WDC}
    \end{figure*}

\begin{figure*}
      \centering
      \setlength{\abovecaptionskip}{3pt}
      \includegraphics[width=\linewidth]{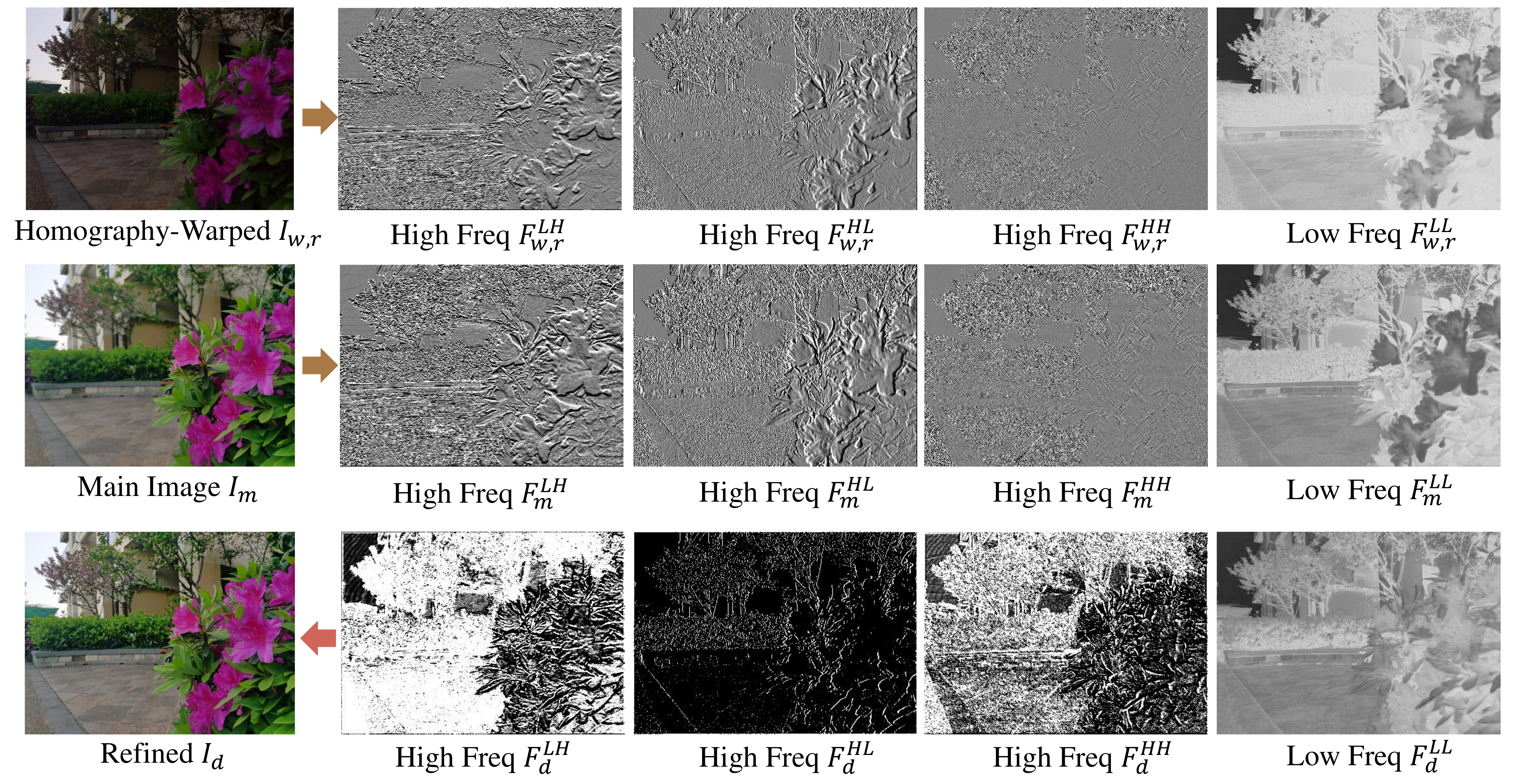}
      \caption{\textbf{Intermediate feature maps of the branch that produces the refined main image in OAS-Net}. The arrows indicate that $I_{w,r}$ and $I_m$ are the inputs, and OAS-Net outputs $I_{d}$ from high frequency features and low frequency features. We can observe that the high frequency feature maps store information of the edges, and the low frequency feature maps relate more to the contents.}
      \label{fig:intermediate2_OAS}
    \end{figure*}

	We show the intermediate results of our approach in~\cref{fig:intermediate1,fig:intermediate2_WDC,fig:intermediate2_OAS}.
    In~\cref{fig:intermediate1}, 
    we show the outputs of each components in EasyAIF, including the flow-warped ultra-wide image $I_{w,f}$, the color-adjusted ultra-wide image $I_{w,c}$, the refined main image $I_d$ and their corresponding fusion masks. In~\cref{fig:intermediate2_WDC,fig:intermediate2_OAS}, we show the high frequency and low frequency feature maps from wavelet transform in WDC-Net and OAS-Net. 

    \begin{figure*}
      \centering
      \includegraphics[width=\linewidth]{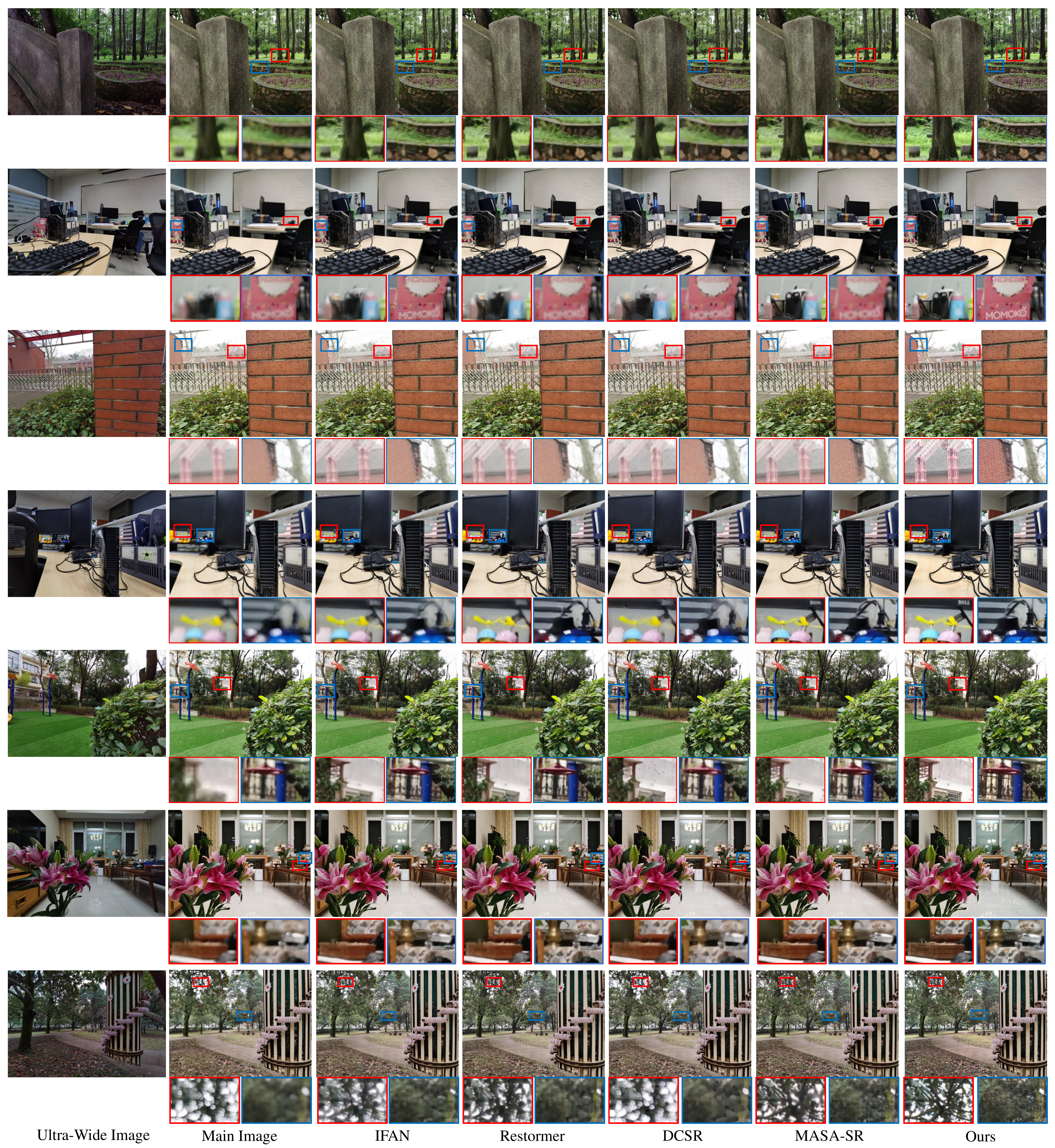}
      \caption{\textbf{Comparison with baselines on the smartphone AIF dataset}. The input pair is the main image focused on the foreground, and the ultra-wide image. Compared with the baselines, our proposed EasyAIF restores sharper details. Please zoom in to see the details.}
      \label{fig:baseline_fg}
    \end{figure*}
    \begin{figure*}
      \centering
      \includegraphics[width=\linewidth]{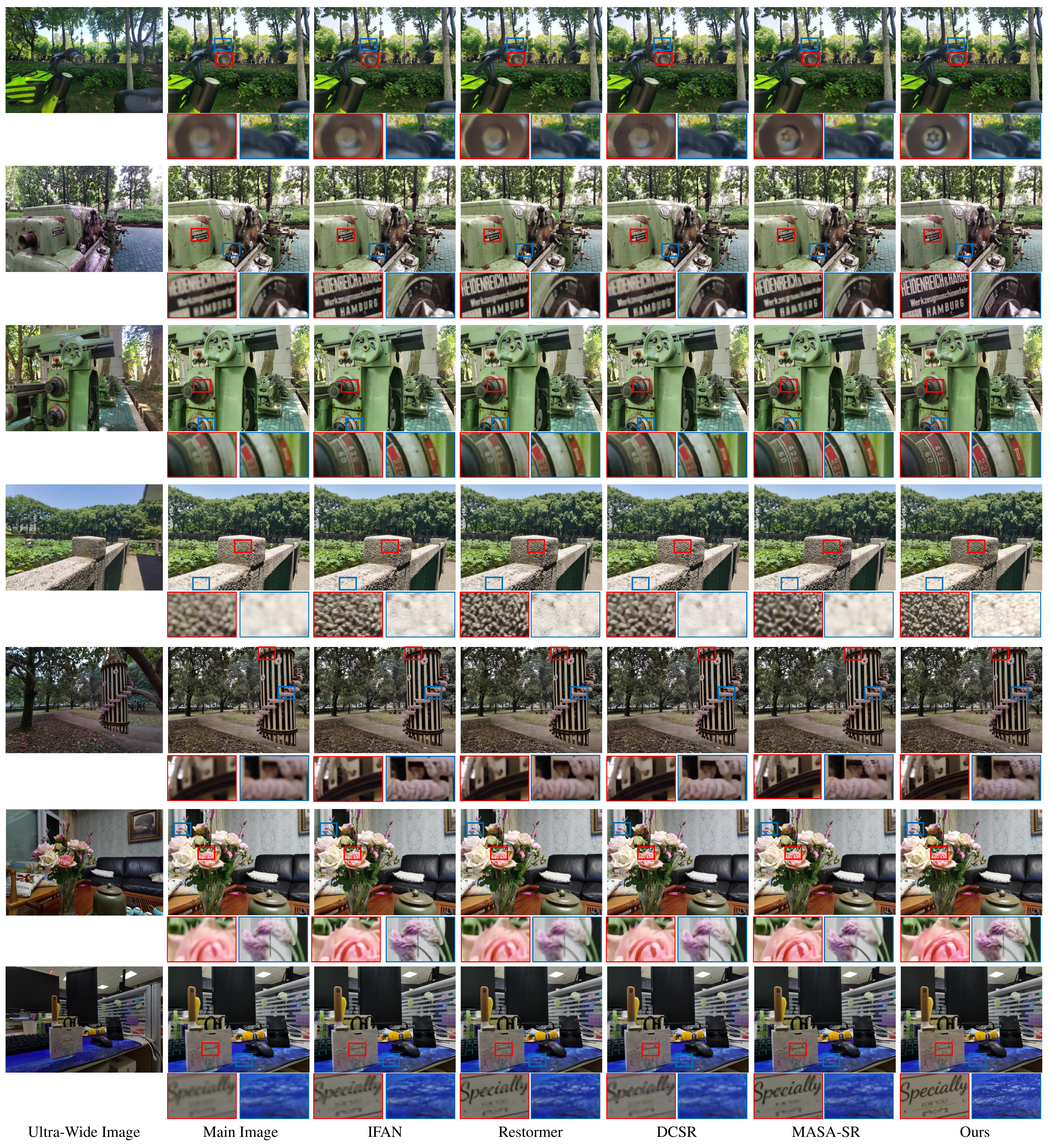}
      \caption{\textbf{Comparison with baselines on the smartphone AIF dataset}. The input pair is the main image focused on the background, and the ultra-wide image. Compared with the baselines, our proposed EasyAIF restores sharper details. Please zoom in to see the details.}
      \label{fig:baseline_bg}
    \end{figure*}

    We show qualitative comparisons with the baseline methods in~\cref{fig:baseline_fg,fig:baseline_bg}. We demonstrate the results with different inputs: the main images focused on the foreground, and the main images focused on the background.
    One can observe: 1) DCSR~\cite{wang2021dual} tends to oversmooth the blurred regions; 2) dual-pixel defocus deblurring methods IFAN~\cite{lee2021iterative} and Restormer~\cite{zamir2021restormer} can effectively reduce the amount of blurring, but they fail to recover sharp details; 3) MASA-SR~\cite{lu2021masa} performs well to restore some sharp details, but it still has many artifacts, such as the blue box in the first row; 4) Our method \ourmethod outperforms other baselines in generating AIF results.

   \subsection{Quantitative Results}

    As shown in Table~\ref{tab:baseline}, we compare our method \ourmethod with the four baseline methods quantitatively on our synthesized smartphone AIF dataset. We split the evaluation dataset into two groups: the main images focused on foreground, and the main images focused on background. To measure the performance of different methods, we use LPIPS~\cite{zhang2018unreasonable}, PSNR, and SSIM as metrics. 
    Results show that \ourmethod outperforms other methods numerically.
    Restormer~\cite{zamir2021restormer} and DCSR~\cite{wang2021dual} are the second best approaches. We demonstrate that \ourmethod is superior in terms of image focused on the foreground as well as on the background. Reference-based super-resolution methods MASA-SR~\cite{lu2021masa} and DCSR perform well to super-resolve the already clear contents in the image, however, they fail to handle large defocus blur. Restormer and IFAN~\cite{lee2021iterative} are designed for defocus deblurring, but they do not fully  utilize the sharp contents in the reference image $I_w$.
    
    To prove the generalization across different devices, we capture 60 scenes on iPhone13 which also has an ultra-wide camera. The resolutions of the ultra-wide image and the main image are both 4032x3024, which are different from those of Huawei Mate30 Pro. We evaluate our model on the iPhone dateset without finetuning. The results are shown in Table~\ref{tab:baseline_iphone}. We show that \ourmethod still performs better than the SOTA SR/Deblur methods.
 
    \begin{table}[t]
        \centering
        \captionsetup{singlelinecheck=false,labelsep=newline,justification=centering}
    	\caption{\scshape Quantitative results on our AIF dataset.
    	The best performance is in \textbf{boldface}, and the second best is \underline{underlined}.} 
    	\resizebox{1.0\linewidth}{!}
    	{
        \setlength{\tabcolsep}{3pt}
    	\renewcommand\arraystretch{1.0}
    	\begin{NiceTabular}{l|ccc|ccc|ccc}
    		\toprule
    		\multicolumn{1}{l}{\multirow{2}{*}[-0.5ex]{Method}} &\multicolumn{3}{c}{Focused on foreground} & \multicolumn{3}{c}{Focused on background}& \multicolumn{3}{c}{Total} \\
    		\cmidrule{2-10}
    		~ &  PSNR$\uparrow$ &  SSIM$\uparrow$&LPIPS$\downarrow$ & PSNR$\uparrow$ & SSIM$\uparrow$ & LPIPS$\downarrow$ &PSNR$\uparrow$ &  SSIM$\uparrow$ & LPIPS$\downarrow$  \\
            \midrule
    		\midrule
    		IFAN~\cite{lee2021iterative} & 25.09 & 0.742 &0.402 &   28.38 & 0.835&0.281 & 26.73& 0.789& 0.341\\
    		Restormer~\cite{zamir2021restormer} & \underline{25.34}& \underline{0.749} & 0.380 &  28.74& 0.840 &0.270 & \underline{27.04}& 0.795 & 0.325\\
    		DCSR~\cite{wang2021dual}  & 24.11 & 0.722 &\underline{0.345} &   \underline{29.66} & \underline{0.875} &\underline{0.203} & 26.88& \underline{0.798} & \underline{0.274}\\
    		MASA-SR~\cite{lu2021masa}& 25.24 & 0.717 &0.422 &   25.72 & 0.722 &0.419& 25.48& 0.719 & 0.421\\
    		\ourmethod (Ours)& \textbf{25.75} & \textbf{0.828 }&\textbf{0.185} & \textbf{30.58}& \textbf{0.921} &\textbf{0.130}& \textbf{28.16}& \textbf{0.874} & \textbf{0.158}\\
    		\bottomrule
    	\end{NiceTabular}
    	}
    	\label{tab:baseline}
    \end{table}
    
    \begin{table}[t]
        \centering
        \captionsetup{singlelinecheck=false,labelsep=newline,justification=centering}
    	\caption{\scshape Quantitative results on iPhone13 dataset.
    	The best performance is in \textbf{boldface}, and the second best is \underline{underlined}.} 
    	\resizebox{1.0\linewidth}{!}
    	{
        \setlength{\tabcolsep}{3pt}
    	\renewcommand\arraystretch{1.0}
    	\begin{NiceTabular}{l|ccc|ccc|ccc}
    		\toprule
    		\multicolumn{1}{l}{\multirow{2}{*}[-0.5ex]{Method}} &\multicolumn{3}{c}{Focused on foreground} & \multicolumn{3}{c}{Focused on background}& \multicolumn{3}{c}{Total} \\
    		\cmidrule{2-10}
    		~ &  PSNR$\uparrow$ &  SSIM$\uparrow$&LPIPS$\downarrow$ & PSNR$\uparrow$ & SSIM$\uparrow$ & LPIPS$\downarrow$ &PSNR$\uparrow$ &  SSIM$\uparrow$ & LPIPS$\downarrow$  \\
            \midrule
    		\midrule
    		IFAN~\cite{lee2021iterative} & \underline{23.48}& \underline{0.638} &0.613 &  26.09 & 0.692&0.527 & 24.78& 0.665& 0.570\\
    		Restormer~\cite{zamir2021restormer} & 23.13& 0.636 & 0.613 &  25.90& 0.690 &0.521 & 24.51& 0.663& 0.567\\
    		DCSR~\cite{wang2021dual}  & 22.78 & \underline{0.638} &\underline{0.479} &   \underline{26.80} & \underline{0.728} &\underline{0.398} & \underline{24.79}& \underline{0.683} & \underline{0.438}\\
    		MASA-SR~\cite{lu2021masa}& 23.09 & 0.590 &0.672 & 23.93 & 0.597 &0.658& 23.51& 0.593 & 0.665\\
    		\ourmethod (Ours)& \textbf{24.60} & \textbf{0.777 }&\textbf{0.241} & \textbf{28.27}& \textbf{0.808} &\textbf{0.187}& \textbf{26.43}& \textbf{0.793} & \textbf{0.214}\\
    		\bottomrule
    	\end{NiceTabular}
    	}
    	\label{tab:baseline_iphone}
    \end{table}

    As shown in Table~\ref{tab:runtime}, we also demonstrate the runtime of \ourmethod. In this table, “Others” means the remaining steps in image registration (descriptor extraction, correspondence matching, and consistency filtering), color alignment, and occlusion-aware synthesis. We evaluate the runtime on the image pair with the resolution of 4032x3024. It is worth mentioning that the SIFT keypoints extraction in image registration is time-consuming, and our total runtime is 1.59s on one NVIDIA 3090 GPU. The optimization of image registration is not our priority, so we simply use SIFT.
    \begin{table}[ht]
    \centering
        \captionsetup{singlelinecheck=false,labelsep=newline,justification=centering}
      \caption{\scshape Runtime of our method. “Others” means the remaining steps in image registration (descriptor extraction, correspondence matching, and consistency filtering), color alignment, and occlusion-aware synthesis.}
      \begin{tabular}{lccc}
        \toprule
        &Keypoints Extraction&  Others &Total\\
        \midrule
        Time (s)& 0.60& 0.99&1.59\\
      \bottomrule
    \end{tabular}
    \label{tab:runtime}
    \end{table}

    \subsection{Ablation Study}
    We conduct ablation study on the framework components, including spatial alignment, color adjustment, and occlusion-aware synthesis.
    \begin{table}[t]
    	\centering
        \captionsetup{singlelinecheck=false,labelsep=newline,justification=centering}
    	\caption{\scshape Ablation study of our AIF synthesis framwork on the smartphone AIF dataset.
    	The best performance is in \textbf{boldface}.} 
    	\resizebox{1.0\linewidth}{!}
    	{
        \setlength{\tabcolsep}{3pt}
    	\renewcommand\arraystretch{1.0}
    	\begin{NiceTabular}{ccc|ccc}
    		\toprule
    		\multicolumn{3}{c}{Framework Components} & \multicolumn{3}{c}{Total} \\
    		\cmidrule{0-5}
    		Spatial alignment& WDC-Net &OAS-Net& PSNR$\uparrow$ &  SSIM$\uparrow$ &LPIPS$\downarrow$  \\
            \midrule
    		\midrule
    		\checkmark &  & &16.06 & 0.675& 0.342\\
    		\checkmark  & \checkmark & & 20.65& 0.732 & 0.304\\
    	      \checkmark   &   &\checkmark & 21.65& 0.739 &0.174 \\
    		 \checkmark  &\checkmark & \checkmark& \textbf{28.16}& \textbf{0.874}& \textbf{0.158}\\
    		\bottomrule
    	\end{NiceTabular}
    	}
    	\label{tab:ablation}
    \end{table}
    \begin{table}
    \centering
        \captionsetup{singlelinecheck=false,labelsep=newline,justification=centering}
      \caption{\scshape Ablation study on warping operations in spatial alignment. The best performance is in \textbf{boldface}.}
      \begin{tabular}{lccc}
        \toprule
        Warping methods&PSNR$\uparrow$ &  SSIM$\uparrow$ & LPIPS$\downarrow$\\
        \midrule
        Homography& 14.37& 0.539&0.431\\
        Flow&15.47&0.584&0.358\\
        Homography$+$Flow &\textbf{16.06 }& \textbf{0.675}& \textbf{0.342}\\
        
      \bottomrule
    \end{tabular}
    \label{tab:spatial_ablation}
    \end{table}
    As shown in Table~\ref{tab:ablation}, our framework works best with all three modules. If we only use spatial alignment, the color discrepancy and the occlusion issue still exist. We mitigate these two problems by introducing WDC-Net and OAS-Net. To further prove our network design, we also conduct experiments on warping in spatial alignment, and ablation study on OAS-Net. As shown in Tab~\ref{tab:spatial_ablation}, homography alignment and flow warping both improve our performance. Homography warping provides image-level alignment, and optical flow focuses on pixel-level warping. 
    \begin{figure}[ht]
      \centering
      \setlength{\abovecaptionskip}{3pt}
        \setlength{\belowcaptionskip}{0pt}
      \includegraphics[width=\linewidth]{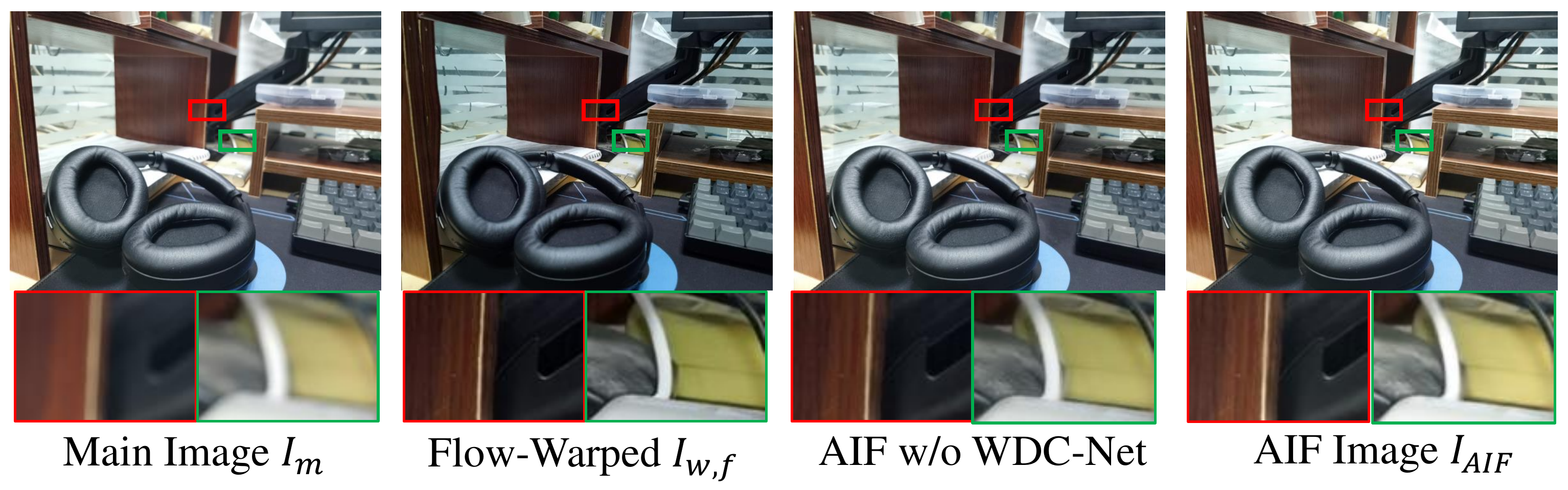}
      \caption{\textbf{Ablation study on our WDC-Net}. From the red and the green boxes, we can observe that before color alignment, flow warped image $I_{w,f}$ is darker than $I_m$, so the AIF result without WDC-Net also exhibits inconsistent color. On the other hand, $I_{AIF}$ shows similar color and illumination to $I_m$. }
      \label{fig:ablation_WDC}
    \end{figure}
    For occlusion-aware synthesis, as shown in Tab~\ref{tab:occlusion_ablation}, we show that wavelet-based deformable refinement, as well as occlusion-aware image fusion, are both required to achieve better quantitative results. In addition, we demonstrate visual qualitative results on the ablation studies of OAS-Net and WDC-Net. 
    In Fig.~\ref{fig:ablation_WDC}, we show that our WDC-Net is effective to align the color of flow-warped image $I_{w,f}$ to the main image $I_m$. For occlusion-aware synthesis, as shown in Fig.~\ref{fig:ablation_OAS}, the color-adjusted $I_{w,c}$ suffers from occlusions, so we use OAS-Net to predict the occluded regions and refine the corresponding areas. Therefore, the final AIF result $I_{AIF}$ performs better than $I_m$ and $I_{w,c}$ on the occlusions.

    \begin{table}[ht]
    \centering
        \captionsetup{singlelinecheck=false,labelsep=newline,justification=centering}
      \caption{\scshape Ablation study on OAS-Net. The best performance is in \textbf{boldface}.}
      \begin{tabular}{cccc}
        \toprule
        OAS structure&PSNR$\uparrow$ &  SSIM$\uparrow$& LPIPS$\downarrow$\\
        \midrule
        Fusion&27.85  & 0.871&\textbf{0.158}\\
        Refine& 26.45& 0.785&0.333\\
        Refine$+$Fusion& \textbf{28.16}& \textbf{0.874}&\textbf{0.158}\\
      \bottomrule
    \end{tabular}
      
    \label{tab:occlusion_ablation}
    \end{table}

    \begin{figure}
      \centering
      \setlength{\abovecaptionskip}{3pt}
        \setlength{\belowcaptionskip}{0pt}
      \includegraphics[width=\linewidth]{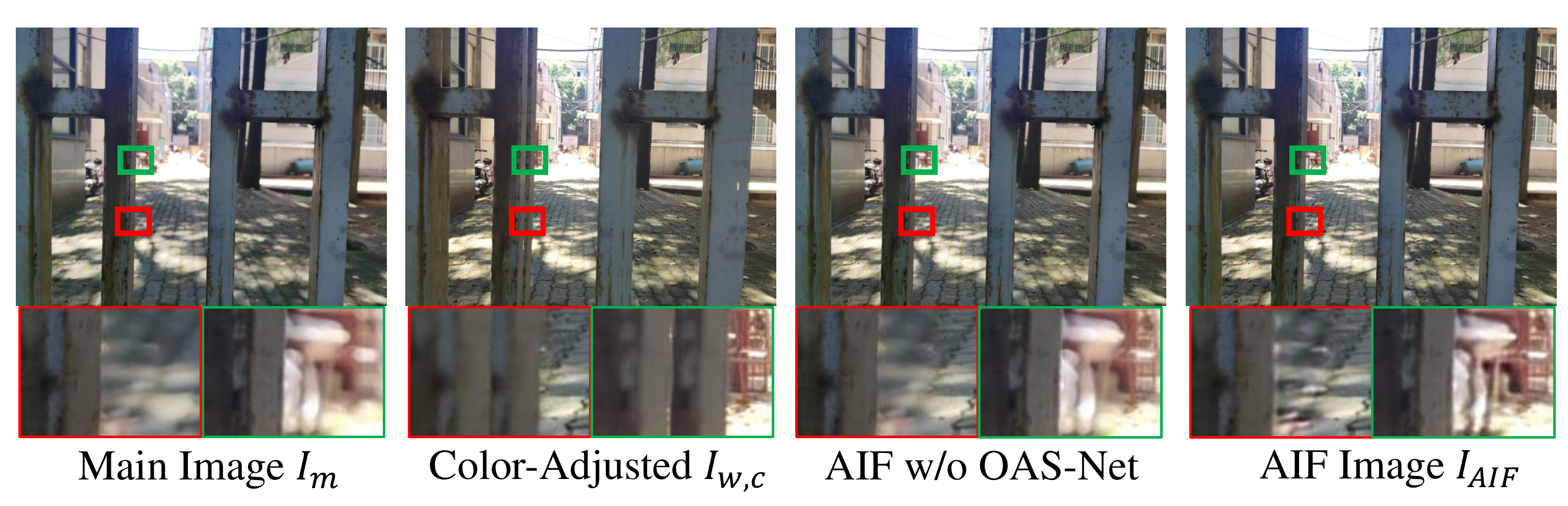}
      \caption{\textbf{Ablation study on our OAS-Net}. From the red and the green boxes, we show that the main image $I_m$ and the color-adjusted $I_{w,c}$ produce blur or artifacts in occluded areas. $I_{AIF}$ has sharper details than the AIF result without OAS-Net, and has fewer artifacts in occlusions than $I_{w,c}$.}
      \label{fig:ablation_OAS}
    \end{figure}
    
    \subsection{User Study}
    To better evaluate the performance of \ourmethod, we conduct a user study on our smartphone AIF dataset. We collect $50$ sets of images, and invite $20$ people participating in this survey. We compare \ourmethod with each baseline separately and ask the participant to choose the more realistic result or choose none if it is hard to judge. 
    
\begin{figure}[htb]
    \begin{center}
    \includegraphics[width=\linewidth]{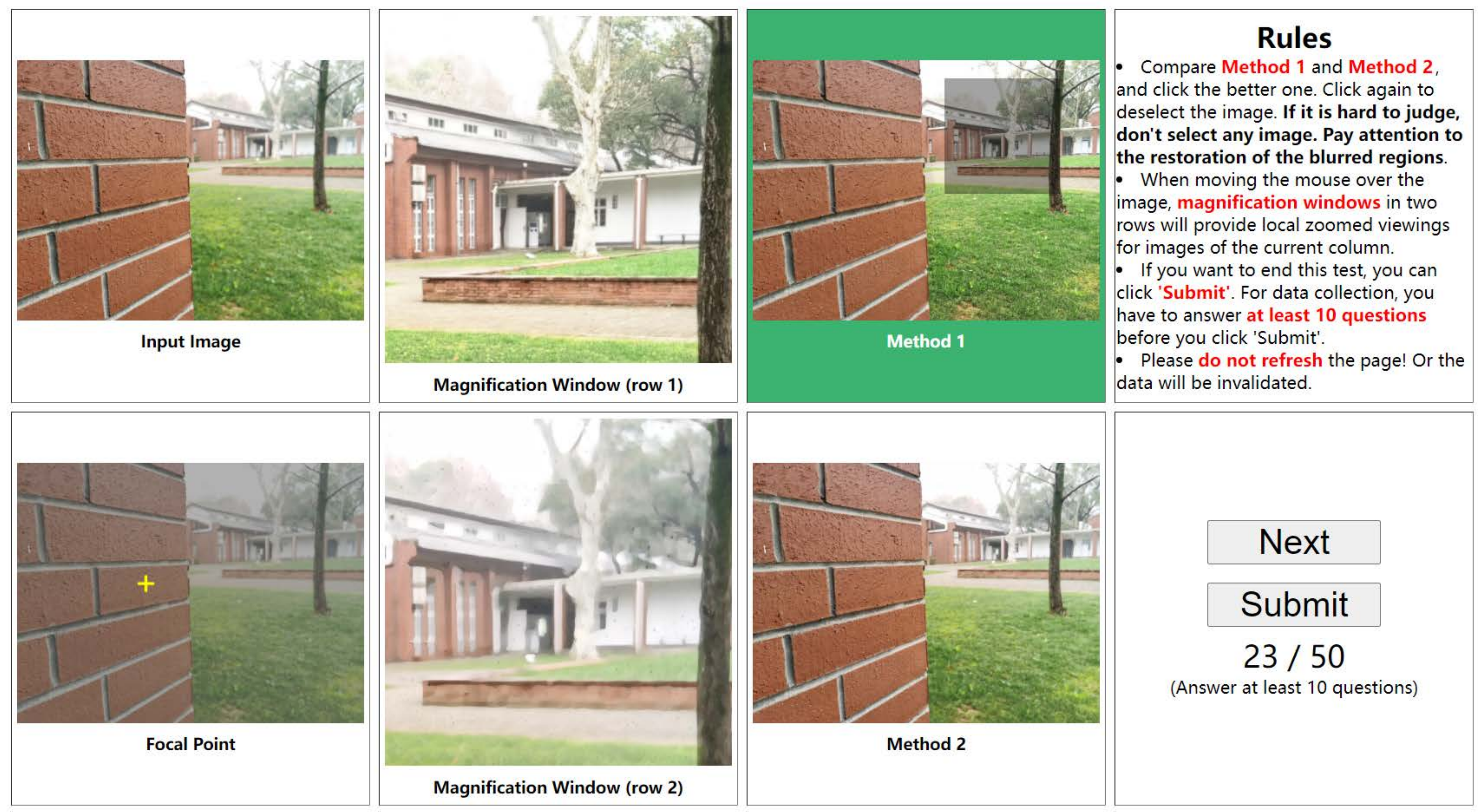}
    \end{center}
    \caption{Interface of the user study website.}
    \label{fig:website}
    \end{figure}

    We build an online website for user study, the interface of the website is demonstrated in Fig.~\ref{fig:website}. ``Input Image'' is a defocused main camera image. ``Focal Point'' labels the target that is roughly refocused during the capturing. ``Method 1'' and ``Method 2'' display the results of two rendering methods, one of which is ours, and the other one is randomly selected from IFAN~\cite{lee2021iterative}, Restormer~\cite{zamir2021restormer}, DCSR~\cite{wang2021dual}, and MASA-SR~\cite{lu2021masa}. The positions of the two methods are also random. ``Magnification Window'' provides simultaneous local zoomed viewings for images in two rows. Users can utilize them to observe and compare the details of the two AIF images. When users are voting, if they cannot decide on the better result, each method gets 0.5 votes.

    We show the comparison results in Table~\ref{tab:userstudy}, where the number represents the preference of our approach over the other methods. 
    One can observe that our approach is most favored.
    \begin{table}[htb]
    	\centering
        \captionsetup{singlelinecheck=false,labelsep=newline,justification=centering}
    	\caption{\scshape User study of \ourmethod on the smartphone AIF dataset. The numbers
        indicate the preference rate of our \ourmethod over other approaches.} 
    	\resizebox{1.0\linewidth}{!}
    	{
        \setlength{\tabcolsep}{3pt}
    	\renewcommand\arraystretch{1.0}
    	\begin{tabular}{ccccc}
        \toprule
        Method&DCSR~\cite{wang2021dual}&MASA-SR~\cite{lu2021masa}&IFAN~\cite{lee2021iterative}&Restormer~\cite{zamir2021restormer} \\
        \midrule
        \ourmethod (Ours)&98.9\% &87.1\% &88.6\% &85.8\%  \\
      \bottomrule
    \end{tabular}
    	}
    	\label{tab:userstudy}
    \end{table}

    \begin{figure}[htb]
      \centering
      \includegraphics[width=\linewidth]{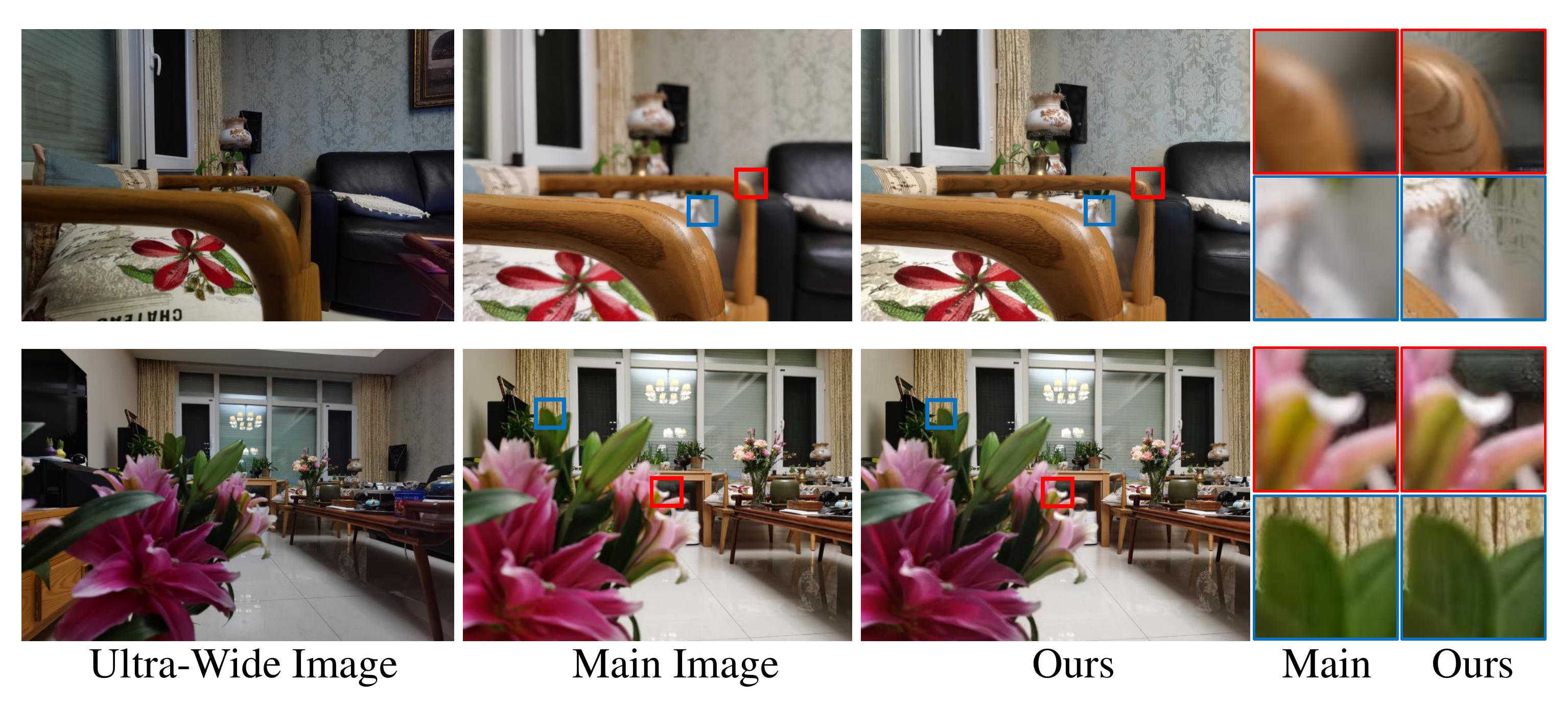}
      \caption{\textbf{Failure cases}. Our All-in-Focus (AIF) result may produce artifacts at edges due to large occlusions (row 1). The ultra-wide image may have defocused regions when the foreground object is too close to the camera, so the corresponding parts of the main image cannot be fixed (row 2).}
      \label{fig:failure}
    \end{figure}

    \subsection{Comparisons with Multi-Focus Fusion Methods}
    \begin{figure}[htb]
      \centering
      \includegraphics[width=\linewidth]{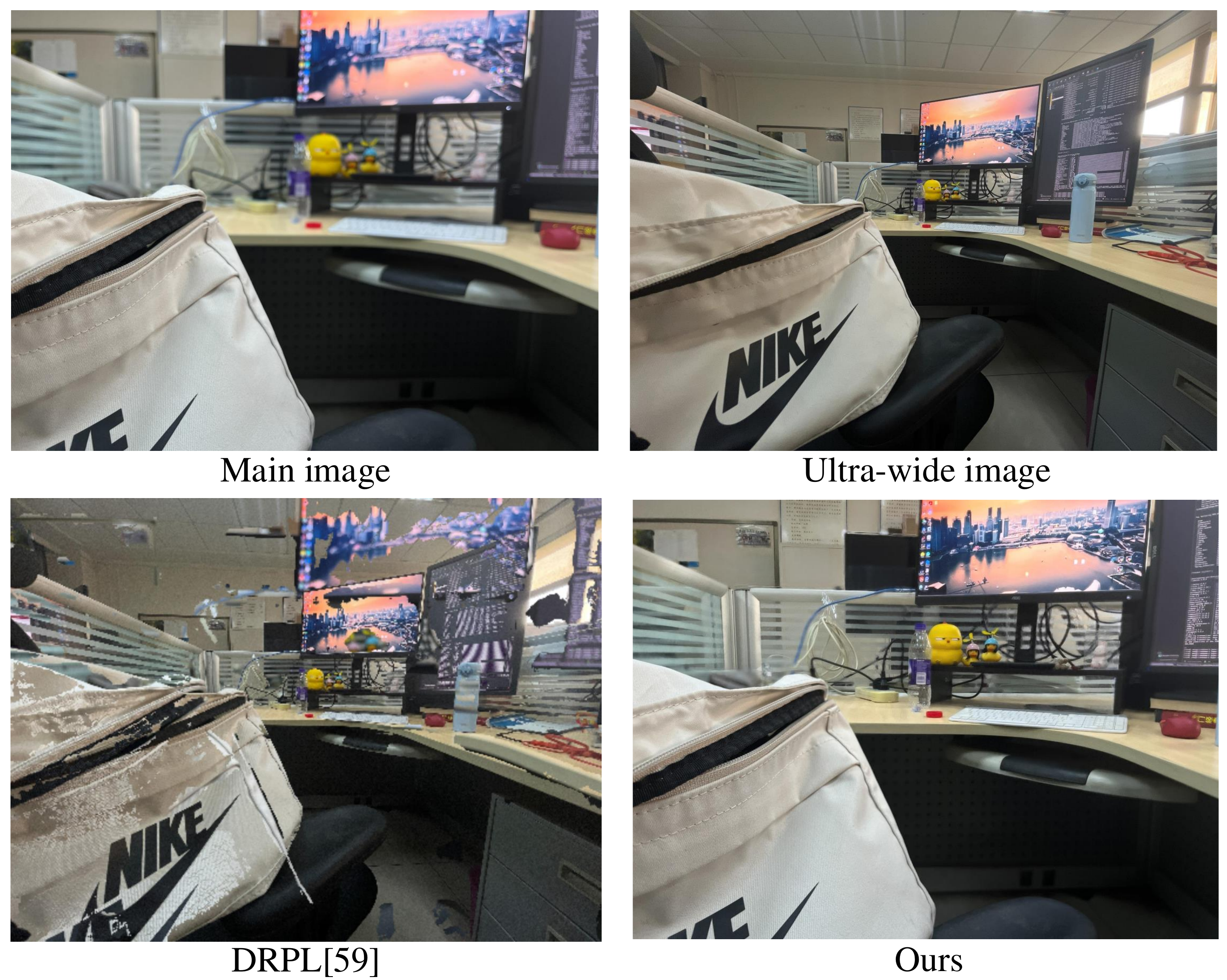}
      \caption{{\textbf{Visualizations of the multi-focus fusion method and \ourmethod on iPhone13 dataset.}. The multi-focus fusion method fails to fuse the two misaligned input images, while \ourmethod is a strong baseline to synthesize an AIF image from main/ultra-wide camera pair.}}
      \label{fig:DRPL}
    \end{figure}
    {Our method is different from multi-focus fusion methods~\cite{zhao2018multi,liu2019new} from the inputs. Our method deals with two images from different cameras, which suffer from large misalignment. However, the input of the multi-focus fusion method is designed to be already aligned. Traditionally the focal stack images are shot by the same camera, so the captured images do not suffer from misalignment in space and color. As shown in Fig.~\ref{fig:DRPL}, multi-focus fusion methods can not deal with the misalignment between the main image and the ultra-wide image.
    In our smartphone AIF task, we use the main camera and the ultra-wide camera from a smartphone, so we need to take misalignment into considerations, which is challenging for our task.
    As shown in Table~\ref{tab:multifocus}, the quantitative result of the multi-focus fusion method is much inferior than the result of \ourmethod.}
    \begin{table}
    \centering
        \captionsetup{singlelinecheck=false,labelsep=newline,justification=centering}
      \caption{{\scshape Comparisons of the multi-focus fusion method and \ourmethod on iPhone13 dataset. The best performance is in \textbf{boldface}.}}
    %   \vspace{-10pt}
      \begin{tabular}{lccc}
        \toprule
        Methods &PSNR$\uparrow$ &  SSIM$\uparrow$ & LPIPS$\downarrow$\\
        \midrule
        DRPL~\cite{li2020drpl}& 11.28& 0.386&0.933\\
        \ourmethod &\textbf{26.43}& \textbf{0.793}& \textbf{0.214}\\
        
      \bottomrule
    \end{tabular}
    \label{tab:multifocus}
    \end{table}
    
	\subsection{Failure Case Analyses}\label{subsec:Failure Cases Analyses}
	
	We show the failure cases in Fig.~\ref{fig:failure}. The occlusion-aware synthesis network OAS-Net may produce artifacts if the occlusions are large. The fusion mask may include the wrongly warped occluded region of the ultra-wide image, and the refined image sometimes cannot fix the blur from large occlusions. In addition, the ultra-wide image may fail to provide sharp details in extreme cases where the camera is too close to the object.

	\section{Conclusion}\label{sec:conclu}
	We are the first to synthesize AIF photos from the main/ultra-wide camera pair. Compared with previous time-consuming methods, we have presented a point-and-shoot solution \ourmethod for AIF photography using a smartphone. 
    We make use of the main/ultra-wide lens pair in modern smartphones to integrate both high-quality details from the main camera and sharp contents from the ultra-wide one. 
    To align the two images, we use spatial warping for spatial alignment, and a wavelet dynamic network for color adjustment. To solve the occlusions brought by parallax, we propose an occlusion-aware synthesis network to refine the occluded regions and predict the fusion mask to generate AIF results. Results show that point-and-shoot AIF photo synthesis is viable from the main and ultra-wide camera pair. 
    Although this framework works well in general, it still has some limitations, such as the inaccurate color transform due to spatial misalignment, and the boundary artifacts caused by imperfect refined results. 
    We will address these issues in our future work.
	
	\bibliographystyle{IEEEtran}
	\bibliography{refs}

\begin{IEEEbiography}[{\includegraphics[width=1in,height=1.25in,clip,keepaspectratio]{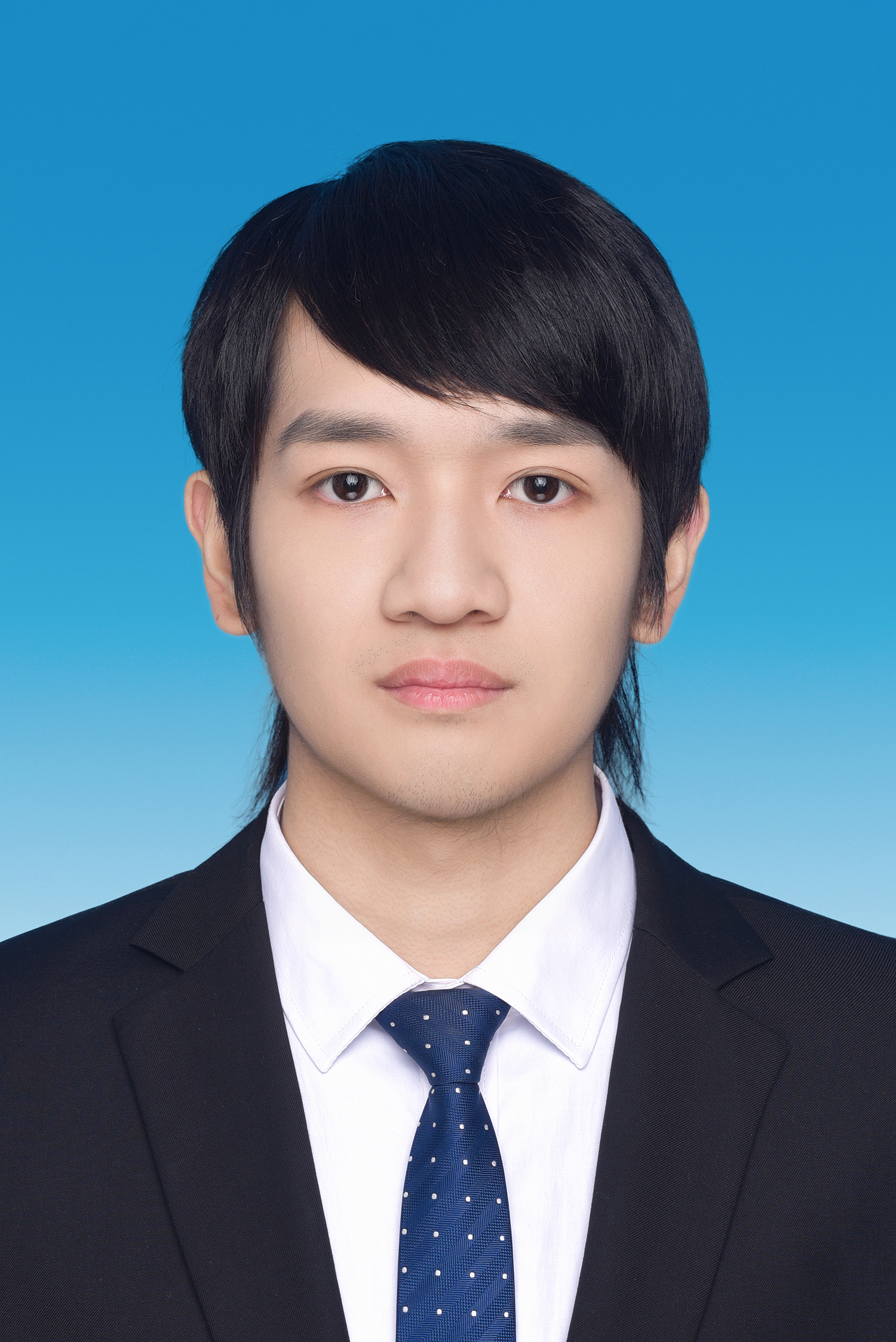}}]{Xianrui Luo}
	received the B.S. degree from Huazhong University of science and Technology, Wuhan, China, in 2020. He is currently pursuing the Ph.D. degree with the School of Artificial Intelligence and Automation, Huazhong University of Science and Technology, Wuhan, China.
	
	His research interests include image restoration and computational photography.
\end{IEEEbiography}

\begin{IEEEbiography}[{\includegraphics[width=1in,height=1.25in,clip,keepaspectratio]{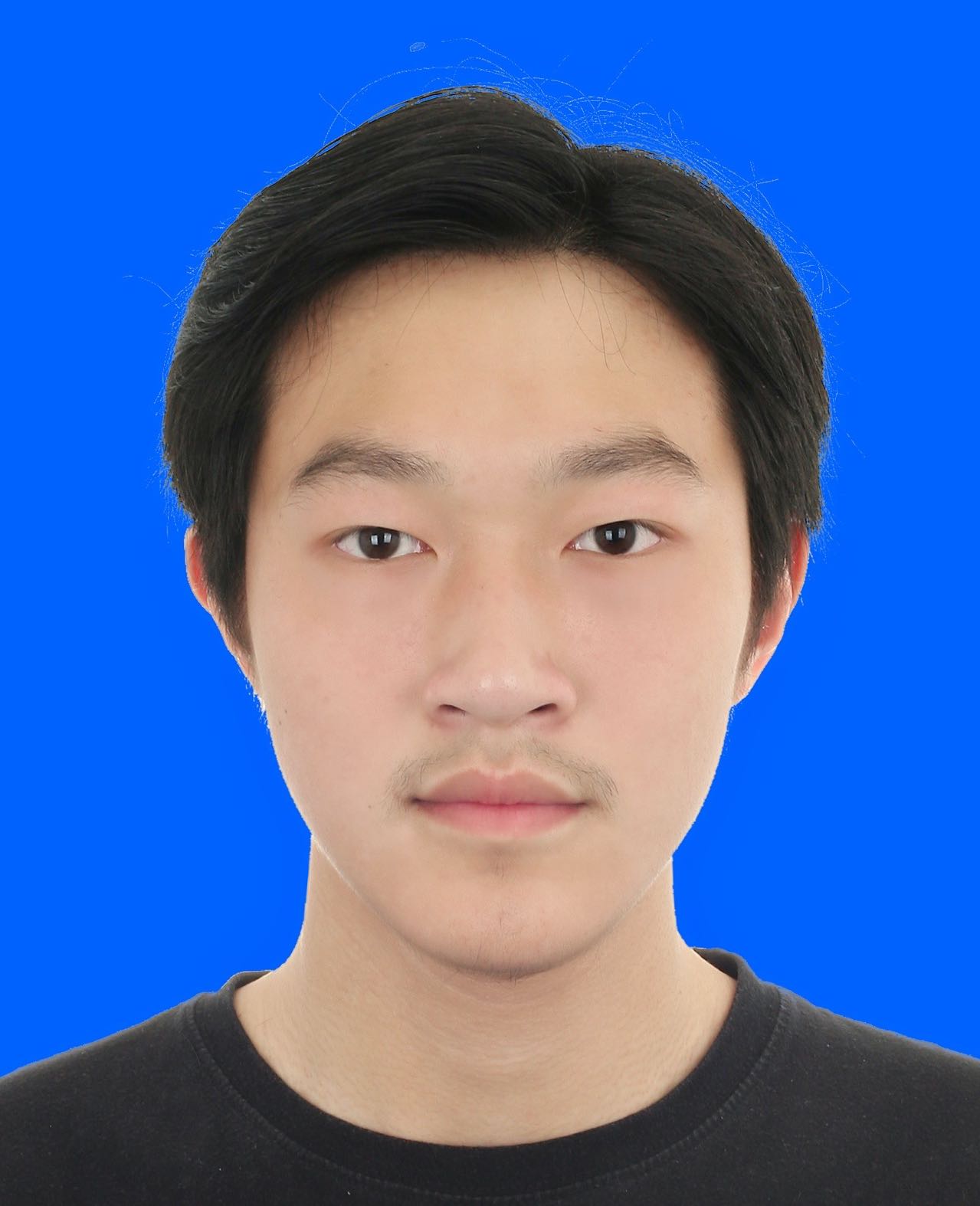}}]{Juewen Peng}
    received the B.S. degree from Huazhong University of science and Technology, Wuhan, China, in 2020. He is currently pursuing the M.S. degree in the School of Artificial Intelligence and Automation, Huazhong University of Science and Technology, Wuhan, China.
     
    His research interests include low-level vision, computational photography, bokeh rendering, deblurring, and image generation.

\end{IEEEbiography}

\begin{IEEEbiography}[{\includegraphics[width=1in,height=1.25in,clip,keepaspectratio]{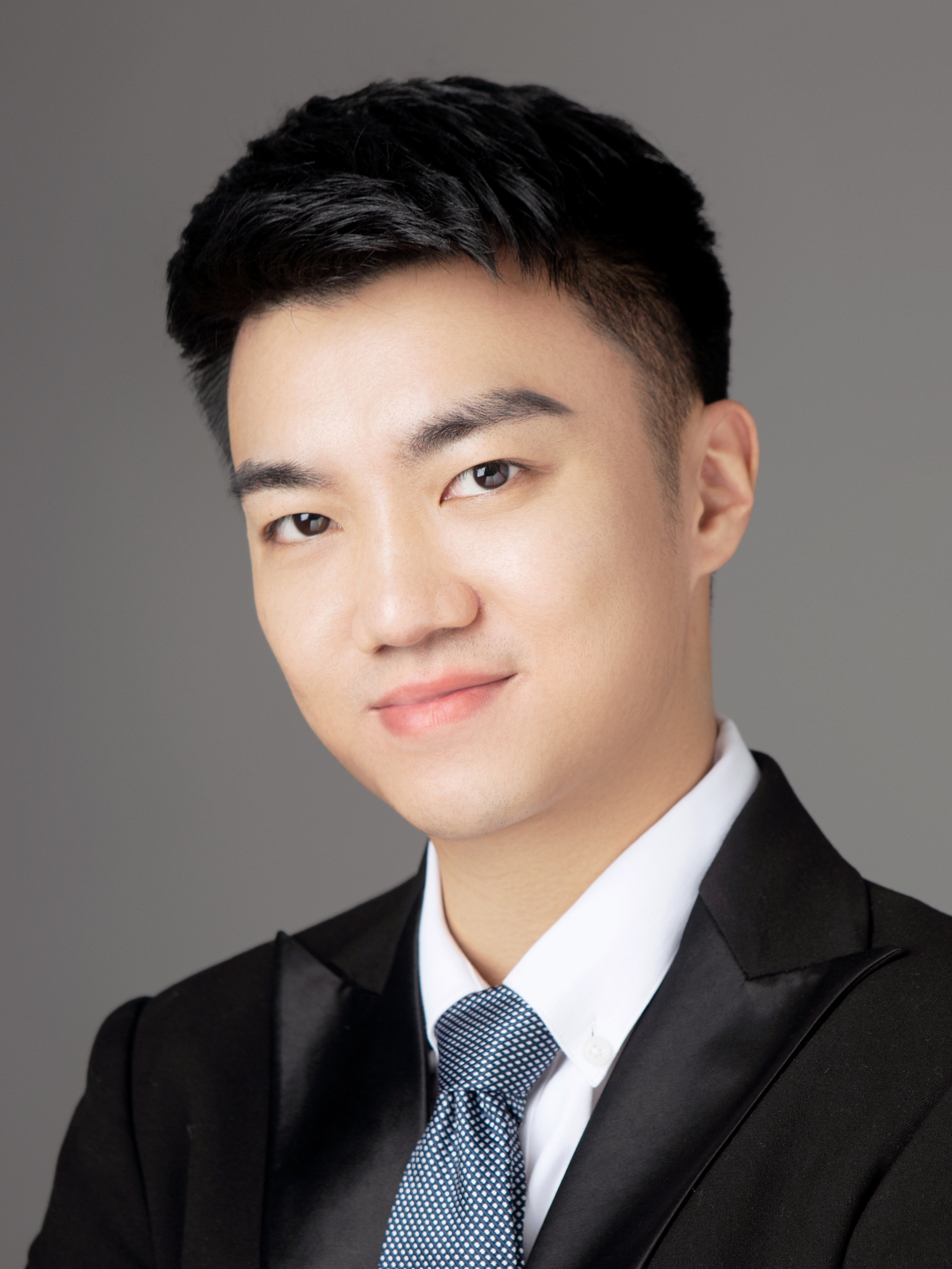}}]{Hao Lu}
    received the Ph.D. degree from Huazhong University of Science and Technology, Wuhan, China, in 2018.
  
  He was a Postdoctoral Fellow with the School of Computer Science, The University of Adelaide, Australia. He is currently an Associate Professor with the School of Artificial Intelligence and Automation, Huazhong University of Science and Technology, China. His research focuses on dense prediction problems in computer vision.
\end{IEEEbiography}

\vskip 0pt plus -1fil

\begin{IEEEbiography}[{\includegraphics[width=1in,height=1.25in,clip,keepaspectratio]{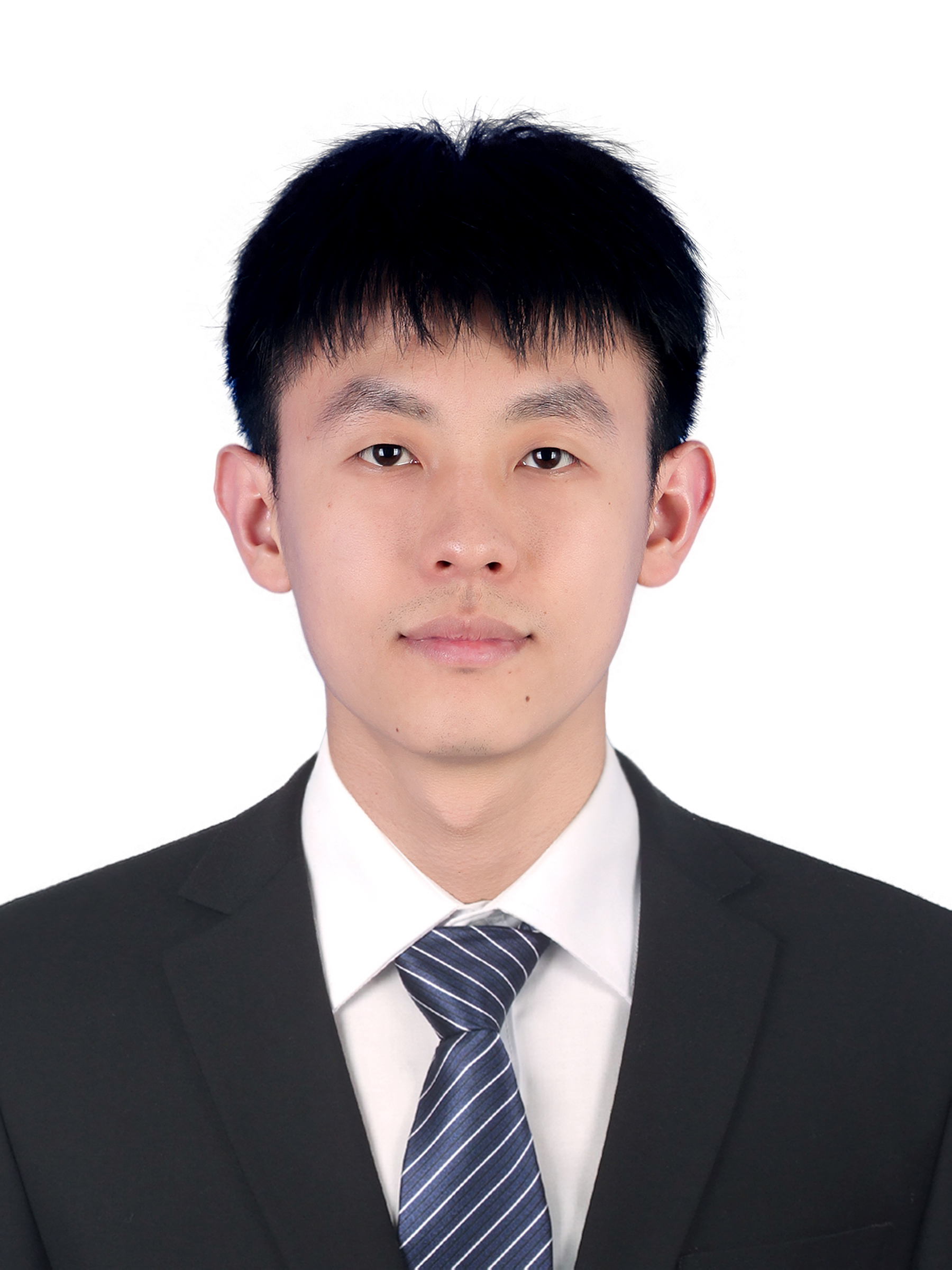}}]{Weiyue Zhao}
	received the B.S. degree from Huazhong University of science and Technology, Wuhan, China, in 2020. He is currently pursuing the M.S. degree with the School of Artificial Intelligence and Automation, Huazhong University of Science and Technology, Wuhan, China.
 
    His research interests include computer vision and machine learning, with particular emphasis on image registration, multi-view stereo and various computer vision applications in video.

\end{IEEEbiography}

\vskip 0pt plus -1fil

\begin{IEEEbiography}[{\includegraphics[width=1in,height=1.25in,clip,keepaspectratio]{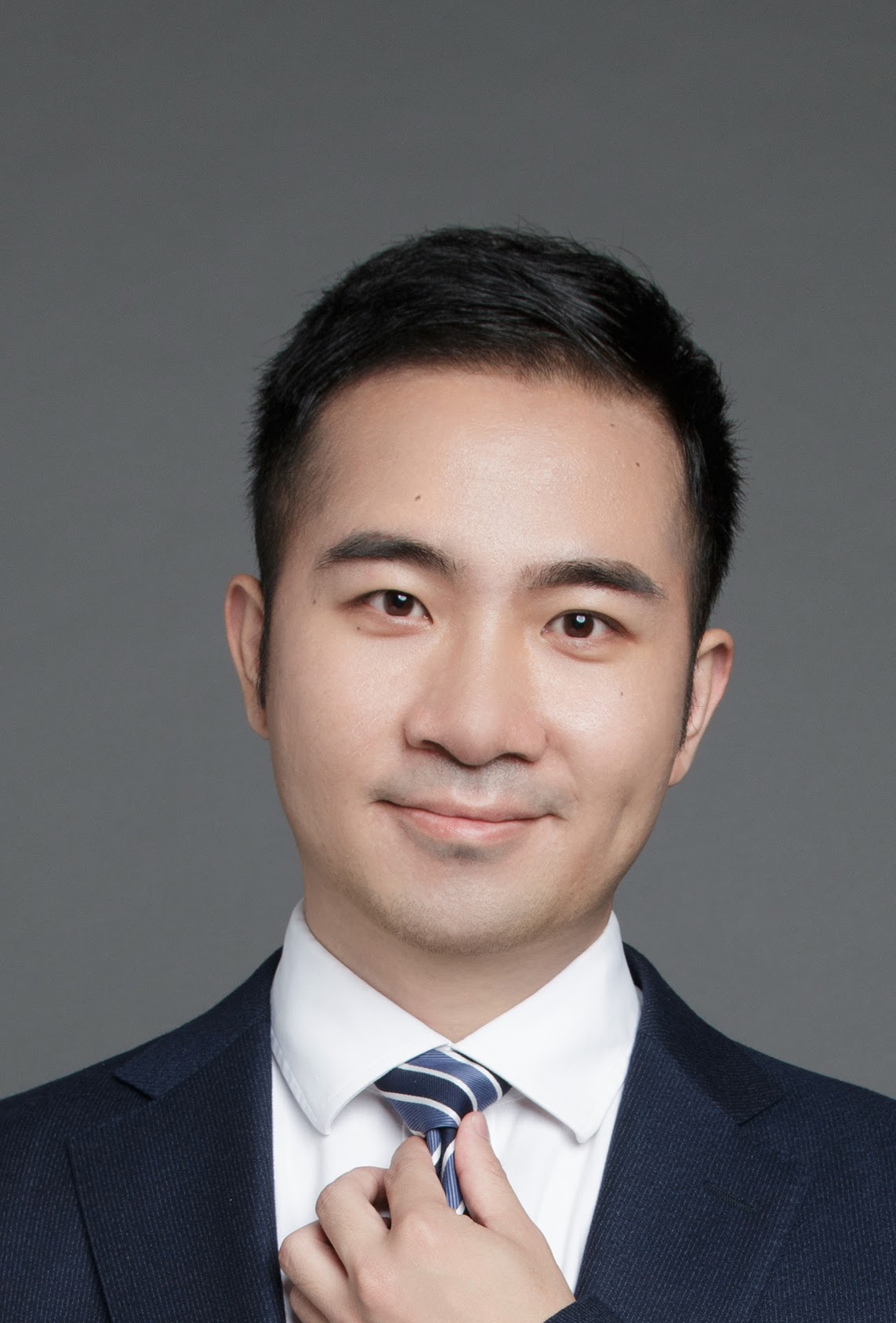}}]{Ke Xian}
	 is a research fellow at S-Lab, Nanyang Technological University (NTU), Singapore. He got his Ph.D. degree at the School of Artificial Intelligence and Automation, Huazhong University of Science and Technology (HUST), China.
	
	His research interests primarily centers on algorithms issues in computer vision and deep learning, including depth estimation from single images, semantic image segmentation and 2D-to-3D conversion.
\end{IEEEbiography}

\vskip 0pt plus -1fil

\begin{IEEEbiography}[{\includegraphics[width=1in,height=1.25in,clip,keepaspectratio]{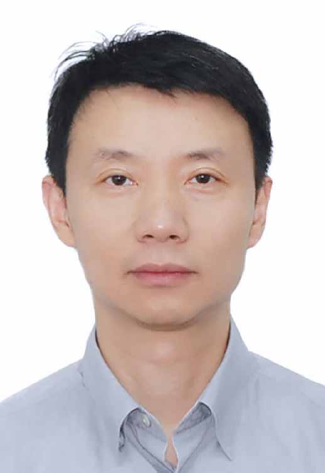}}]{Zhiguo Cao} (Member, IEEE) received the B.S. and M.S. degrees in communication and information system from the University of Electronic Science and Technology of China and the Ph.D. degree in pattern recognition and intelligent system from Huazhong University of Science and Technology. 

He is currently a Professor with the School of Artificial Intelligence and Automation, Huazhong University of Science and Technology. His research interests spread across image understanding and analysis, depth information extraction, 3d video processing, motion detection, and human action analysis. He has published dozens of papers at international journals and prominent conferences, which have been applied to automatic observation system for crop growth in agricultural, for weather phenomenon in meteorology and for object recognition in video surveillance system based on computer vision.
\end{IEEEbiography}

\vskip 0pt plus -1fil

\vfill

\end{document}